%% file: dong2022ras.tex
\documentclass[a4paper,fleqn]{cas-dc}

\usepackage{lineno,hyperref}
\modulolinenumbers[5]

\input{stachnisslab-latex}

\input{stachnisslab-math}

\usepackage{color}
\usepackage{multicol}
\usepackage{siunitx}
\usepackage{subcaption}
\usepackage{graphics}    
\usepackage{times}       
\usepackage{amsmath}     
\usepackage{amssymb}     
\usepackage{graphicx}
\usepackage{xcolor}
\usepackage[algo2e, linesnumbered,ruled,noline]{algorithm2e} 
\usepackage{multirow}

\DeclareMathOperator{\RMSE}{RMSE}

\renewcommand{\v}[1]{{\mathbf{#1}}}

\SetKwInput{KwInput}{Input}                
\SetKwInput{KwOutput}{Output} 

\AtBeginEnvironment{algorithm}{\SetArgSty{textrm}}

\usepackage[numbers]{natbib}

\begin{document}
\let\WriteBookmarks\relax
\def\floatpagepagefraction{1}
\def\textpagefraction{.001}

\shorttitle{Online Pole Segmentation on Range Images for Long-term LiDAR Localization in Urban Environments}    

\shortauthors{H. Dong, X. Chen, S. S\"arkk\"a, C. Stachniss}  

\title[mode = title]{Online Pole Segmentation on Range Images \\ for Long-term LiDAR Localization in Urban Environments}

\author[]{Hao~Dong$^*$~\quad~Xieyuanli~Chen$^*$\textdagger~\quad~Simo~S\"arkk\"a~\quad~Cyrill~Stachniss}
\nonumnote{$^*$Authors with equal contribution.}
\nonumnote{\textdagger~Corresponding author: xieyuanli.chen@igg.uni-bonn.de}
\nonumnote{H. Dong and S. S\"arkk\"a are with the Aalto University, Finland. X. Chen and C. Stachniss are with the University of Bonn, Germany. C. Stachniss is additionally with the Department of Engineering Science at the University of Oxford, UK, and with the Lamarr Institute for Machine Learning and Artificial Intelligence, Germany. This work has been funded by the Deutsche Forschungsgemeinschaft (DFG, German Research Foundation) under Germany's Excellence Strategy, EXC-2070 -- 390732324 (PhenoRob).}

\begin{abstract}
Robust and accurate localization is a basic requirement for mobile autonomous systems. Pole-like objects, such as traffic signs, poles, and lamps are frequently used landmarks for localization in urban environments due to their local distinctiveness and long-term stability. In this paper, we present a novel, accurate, and fast pole extraction approach based on geometric features that runs online and has little computational demands. Our method performs all computations directly on range images generated from 3D LiDAR scans, which avoids processing 3D point clouds explicitly and enables fast pole extraction for each scan. We further use the extracted poles as pseudo labels to train a deep neural network for online range image-based pole segmentation.
We test both our geometric and learning-based pole extraction methods for localization on different datasets with different LiDAR scanners, routes, and seasonal changes. The experimental results show that our methods outperform other state-of-the-art approaches. 
Moreover, boosted with pseudo pole labels extracted from multiple datasets, our learning-based method can run across different datasets and achieve even better localization results compared to our geometry-based method. 
We released our pole datasets to the public for evaluating the performance of pole extractors, as well as the implementation of our approach.
\end{abstract}

\begin{keywords}
\texttt{Localization \sep Pole \sep LiDAR \sep Range image \sep Mapping \sep Autonomous driving \sep Deep learning \sep Semantic segmentation}
\end{keywords}

\maketitle


\section{Introduction}
\label{sec:intro}
Robust and precise localization is a crucial capability for an autonomous robot and a commonly performed state estimation task~\cite{10.5555/3165227}. 
The accurate estimation of the robot's pose helps to avoid collisions, navigate in a goal-directed manner, follow the traffic lanes, and perform other tasks. 
Reliability here means that the robot should adapt to changes in the environment, such as different weather conditions~\cite{carlevaris-bianco2016ijrr}, day and night~\cite{toft2020tpami}, or seasonal changes~\cite{RobotCar}.

Global navigation satellite system-based localization systems are robust to appearance changes of the environments. 
However, in urban areas, they may suffer from low availability due to building and tree occlusions. 
Additional, map-based approaches are needed for precise and reliable localization for mobile robots. 
Multiple different types of sensors have been used to build the map of the environments, including Light Detection and Ranging (LiDAR) scanners~\cite{droeschel2018icra, behley2018rss, vizzo2021icra}, monocular~\cite{mur-artal2015tro} and stereo cameras~\cite{cvisic2017jfrnewbib}. 
Among them, LiDAR sensors are more robust to the illumination changes, and multiple LiDAR-based effective and efficient mapping approaches have been proposed, for example, by Behley and Stachniss~\cite{behley2018rss} or by Droeschel and Behnke~\cite{droeschel2018icra}.
However, these approaches often need substantial amounts of memory due to map representations, thus cannot generalize easily to large-scale scenes. 
If only specific features are used to build the map, such as traffic signs, trunks and other pole-like structures, the map size can be reduced significantly~\cite{wilbers2021phd}.

\begin{figure}[t]
	\centering
	\begin{subfigure}[b]{\linewidth}
	  \includegraphics[width=\linewidth]{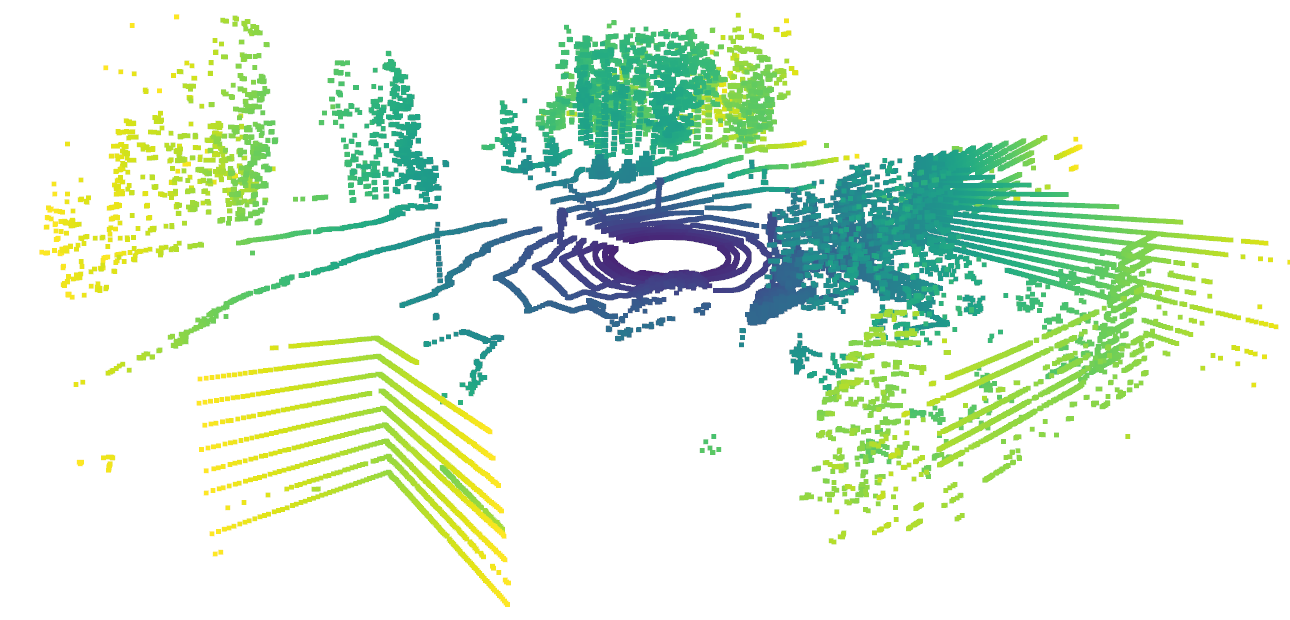}
	  \caption{Current scan}
	\end{subfigure}
	\begin{subfigure}[b]{\linewidth}
	  \vspace{1mm}
	  \includegraphics[width=\linewidth]{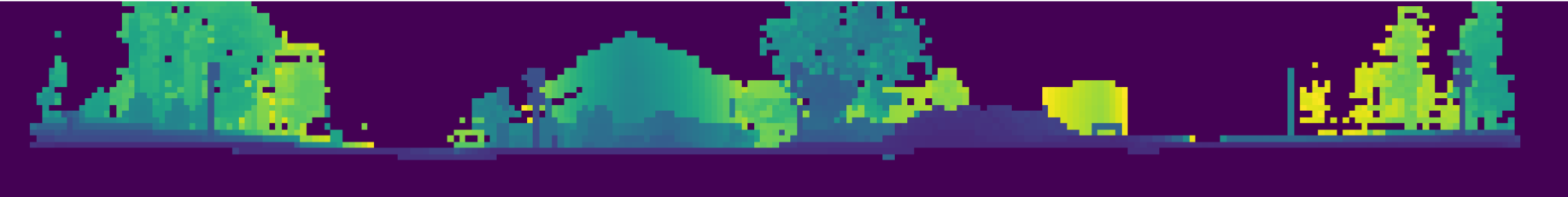}
	  \caption{Current range image}
	\end{subfigure}
	\begin{subfigure}[b]{\linewidth}
	  \vspace{1mm}
		\includegraphics[width=\linewidth]{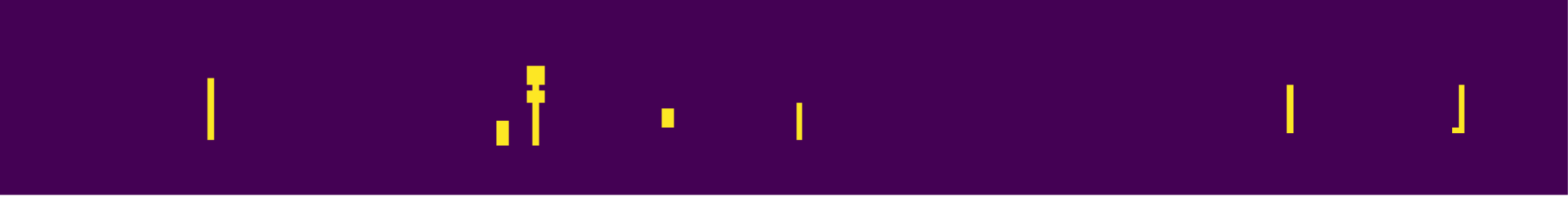}
		\caption{Pole extraction result by our approach}
	  \end{subfigure}
	\caption{Visualization of range image and pole extraction. On the top is the raw LiDAR scan. The corresponding range image generated from this scan is in the middle. 
	The bottom is the pole extraction result based on the range image.}
	\label{fig:pole_extractor}
  \end{figure}

The main contribution of this paper is a novel range image-based pole extractor that can be used for long-term localization of autonomous mobile systems. 
Instead of using the raw point clouds obtained from 3D LiDAR sensors directly, we investigate the use of range images for pole extraction. 
Range image is a light and natural representation of the scan from a rotating 3D LiDAR such as a Velodyne or Ouster sensors. 
Operating on such an image is considerably faster than on the raw 3D point cloud. 
Besides, a range image keeps the neighborhood information implicitly in its 2D structure and we can use this information for segmentation.
The detected poles in the range image can further be used as pseudo pole labels to train a pole segmentation neural network.
After training once with pseudo pole labels generated from different datasets, our learning-based based method can detect poles in different environments and achieve even better localization performance than our geometry-based method. 
To achieve LiDAR localization, in the mapping phase, we first project the raw point cloud into a range image and then extract poles from that image,  as shown in \figref{fig:pole_extractor}. 
After obtaining the position of poles in the range image, we use the ground-truth poses of the robot to reproject them into the global coordinate system to build a global map. 
During localization, we utilize Monte Carlo localization~(MCL) for updating the importance weights of the particles by matching the poles detected from online sensor data with the poles in the global map. 

In sum, we make three key claims that our approach is able to
(i)~extract more reliable poles in the scene compared to the baseline method, as a result, 
(ii)~achieve better online localization performance in different environments, and 
(iii)~generate pseudo pole labels to train a pole segmentation network achieving better localization results and faster runtime compared to the geometric method.
These claims are backed up by the paper and our experimental evaluation.
The code of our approach and the pole dataset are released at:
\href{https://github.com/PRBonn/pole-localization}{https://github.com/PRBonn/pole-localization}.

\section{Related Work}
\label{sec:related}

For localization given a map, there exists a large amount of research.
While many different types of sensors have been used to tackle this problem~\cite{thrun2005probrobbook}, in this work, we mainly concentrate on LiDAR-based approaches.

Traditional approaches to robot localization rely on probabilistic state estimation techniques. 
A popular framework is Monte Carlo localization~\cite{dellaert1999icra}, which uses a particle filter to estimate the pose of the robot and is still widely used in robot localization systems~\cite{bennewitz2006euros,chen2020iros,chen2021icra,kummerle14jfr,sun2020icra,trahanias2005ram,zimmerman2022iros}.

Besides the traditional geometry-based methods, more and more approaches recently exploit deep neural networks and semantic information for 3D LiDAR localization. 
For example, Ma \etal~\cite{ma2019iros} combine semantic information such as lanes and traffic signs in a Bayesian filtering framework to achieve accurate and robust localization within sparse HD maps, whereas Tinchev \etal~\cite{tinchev2019ral} propose a learning-based method to match segments of trees and localize in both urban and natural environments.
Sun \etal~\cite{sun2020icra} use a deep-probabilistic model to accelerate the initialization of the Monte Carlo localization and achieve a fast localization in outdoor environments.
Shi \etal~\cite{shi2021ral} exploit a graph-based network to register LiDAR point clouds.
Wiesmann \etal~\cite{wiesmann2021ral} propose a deep learning-based 3D network to compress the LiDAR point cloud, which can be used for large-scale LiDAR localization.
In our previous work~\cite{chen2021auro,chen2020rss,chen2020iros}, we also exploit CNNs with semantics to predict the overlap between LiDAR scans as well as their yaw angle offset and use this information to build a learning-based observation model for Monte Carlo localization.
The learning-based methods perform well in the trained environments, while they usually cannot generalize well in different environments or different LiDAR sensors.

Instead of using dense semantic information estimated by neural networks~\cite{milioto2019iros,milioto2019icra,li2022ral,cortinhal2020iv}, a rather lighter solution has been proposed for long-term localization, which extracts only pole landmarks from point clouds.
There are usually two parts in pole-based approaches, pole extraction and pose estimation.
For pole extraction, Sefati \etal~\cite{sefati2017iv} first remove the ground plane from the point cloud and project the rest points on a horizontal grid. 
After that, they cluster the grid cells and fit a cylinder for each cluster. 
Finally, a particle filter with nearest-neighbor data association is used for pose estimation.
Weng \etal~\cite{weng2018rcar} and Schaefer \etal~\cite{schaefer2019ecmr} use similar particle filter-based methods to estimate the pose of the robot with different pole extractors. 
Weng \etal~\cite{weng2018rcar} discretize the space and extract poles based on the number of laser reflections in each voxel. 
Based on that, Schaefer \etal~\cite{schaefer2019ecmr} consider both the starting and end points of the scan and thus model the occupied and free space explicitly.
K{\"u}mmerle~\etal~\cite{kummerle2019icra} use a nonlinear least-squares optimization method to refine the pose estimation. Spangenberg~\etal~\cite{spangenberg2016iros} use stereo camera images to extract poles and then feed them into a particle filter with odometry and GPS data. Shi~\etal~\cite{rs10121891} extract pole-like objects from the point cloud by spatial independence analysis and cylindrical or linear feature detection. They also classify the pole-like objects into street lamps, traffic signs and utility poles by 3D shape matching. Weng~\etal~\cite{weng2016road} exploit the reflective intensity information to extract traffic signs which are always painted with highly reflective materials. Chen~\etal~\cite{9548787} fuse poles information into a non-linear optimization problem to obtain the vehicle location. Plachetka~\etal~\cite{9564759} use a deep neural network for pole extraction by learning encodings of the point cloud input.
In contrast to the aforementioned approaches, we use a projection-based method and avoid the comparable costly processing of 3D point cloud data. Thus, our implementation is fast.

This article is an extension of our previous conference paper~\cite{dong2021ecmr}. In our previous work, we propose a geometry-based pole extractor on LiDAR point clouds, which uses only range information without exploiting neural networks or deep learning. Thus, it generalizes well to different environments and different LiDAR sensors and does not require new training data when moving to different environments. 
Inspired by an automatically labeling method~\cite{chen2022ral}, in this article, we further use the poles extracted by our geometry-based method as pseudo labels to train a pole segmentation network.
Trained with a large number of pseudo pole labels automatically generated by our geometry-based pole extractor on different datasets, our learning-based method can generalize well in different environments and outperforms the geometry-based method.
As for the network architecture, we use SalsaNext~\cite{cortinhal2020iv}, a state-of-the-art range image-based semantic segmentation network. 
Instead of segmenting the environment into multiple classes like ground, structure, vehicle and human, in our case, we only distinguish the poles from other objects.

\begin{figure}[t]
	\includegraphics[width=\linewidth]{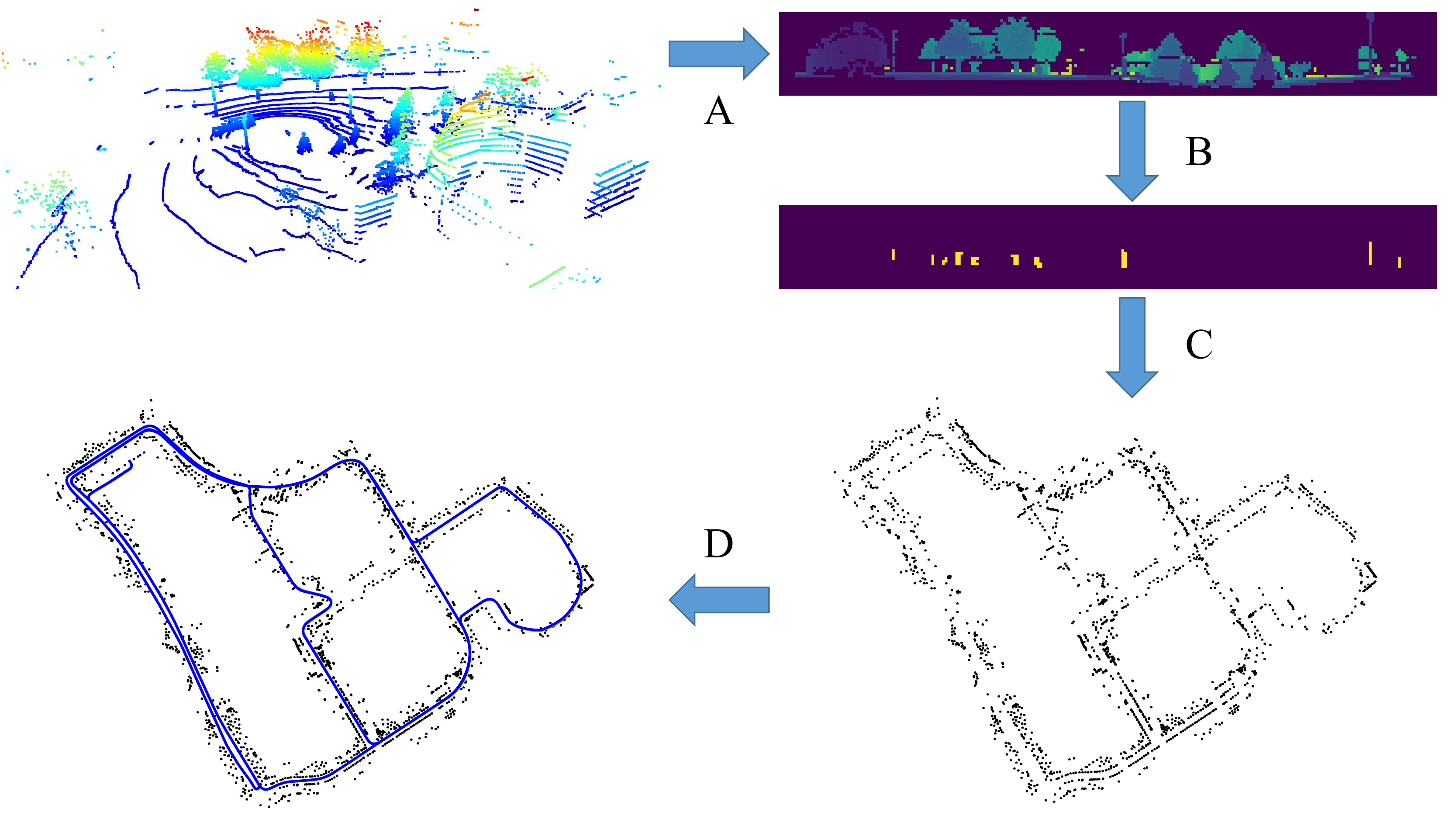}
	\caption{Overview of our pole-based localization system. A, we project the LiDAR point cloud into a range image, and B, extract poles in the image. C, based on the extracted poles, we then build a global pole map of the environment. D, we finally propose a pole-based observation model for MCL to localize the robot in the map.}
	\label{fig:system}
  \end{figure}

\section{Methodology}
\label{sec:main}

In this paper, we propose a range image-based pole extractor for long-term localization using a 3D LiDAR sensor.
As shown in \figref{fig:system}, we first project the LiDAR point cloud into a range image (\secref{rangeGe}) and extract poles from it using either a geometric (\secref{poleEx}) or a learning-based (\secref{poleExLearning}) method. 
Based on the proposed pole extractor, we then build a global pole map of the environment (\secref{map}).
In the localization phase, we extract poles online using the same extractor and use a novel pole-based observation model for Monte Carlo localization (\secref{loc}).

\subsection{Range Image Generation\label{rangeGe}}
The key idea of our approach is to use range images generated from LiDAR scans for pole extraction. 
Following the prior work~\cite{chen2019iros, chen2021icra}, we utilize a spherical projection for range images generation. 
Each LiDAR point $\v{p}=(x, y, z)$ is mapped to spherical coordinates via a mapping $\Pi: \RR^3 \mapsto \RR^2$ 
and finally to image coordinates, as defined by
\begin{align}
	\left( \begin{array}{c} u \vspace{0.0em} \\ v \end{array}\right) & = \left(\begin{array}{cc} \frac{1}{2}\left[1-\arctan(y, x) \, \pi^{-1}\right]~\,~w   \vspace{0.5em} \\
			\left[1 - \left(\arcsin(z\, r^{-1}) + \abs{\mathrm{f}_{\mathrm{down}}}\right) \mathrm{f}^{-1}\right] \, h\end{array} \right), \label{eq:projection}
\end{align}
where $(u,v)$ are image coordinates, $(h, w)$ are the height and width of the desired range image, $\mathrm{f}~{=}~\abs{\mathrm{f}_{\mathrm{up}}}~{+}~\abs{\mathrm{f}_{\mathrm{down}}}$ is the vertical field-of-view (FOV) of the sensor ($\mathrm{f}_{\mathrm{up}}$ is the up vertical FOV and $\mathrm{f}_{\mathrm{down}}$ is the down vertical FOV). The $\mathrm{f}_{\mathrm{up}}$ is a positive value in radian, while $\mathrm{f}_{\mathrm{down}}$ is a negative value in radian. The $r~{=}~||\v{p}_i||_2$ is the range value of each point.
This procedure results in a list of~$(u,v)$ tuples containing a pair of image coordinates for each $\v{p}_i$, which we use to generate our proxy representation. Using these indices, we extract for each $\v{p}_i$, its range~$r$, its $x$, $y$, and~$z$ coordinates, and store them in the image.

\subsection{Geometry-based Pole Extractor}
\label{poleEx}

\begin{algorithm2e}[t]

	\small
    \SetAlgoLined
	\DontPrintSemicolon
	\KwInput{Range Image $\mathbf{I}_{range}$}
	\KwOutput{PoleParameters $\mathbf{P}$ with circle centers and radiuses}
	Let $\mathbf{O}$ be the set of all valid pixel coordinates in $\mathbf{I}_{\mathit{range}}$. $T_d$ is the distance threshold to find neighbors, $T_n$ is the pixel count threshold, and $T_h$ is the object height threshold. $H_a$ and $H_b$ are the height lower bound and higher bound. $R_a$ and $R_b$ are the radius lower bound and higher bound.\;
	\While{$\mathbf{O} \neq \emptyset$ }
	{	create a new $\mathbf{c}$ in $\mathbf{C}$;
	    $\mathbf{p} \gets \mathbf{O}[0]$;
	    $\mathbf{O} \gets \mathbf{O} - \mathbf{p}$;
	    $\mathbf{c} \gets \mathbf{c} + \{\mathbf{p}\}$; 
		create an empty $\mathbf{N}$ \;
		\ForEach{$\mathbf{p}^{'} = Neighbor(\mathbf{p})  \in \mathbf{O}$ \emph{and} $Distance(\mathbf{p}^{'}, \mathbf{p}) \textless T_d$}
			{
				$\mathbf{N} \gets \mathbf{N} + \mathbf{p}^{'}$ \;	
			}
		\While{$\mathbf{N} \neq \emptyset$}
		{   
		    $\mathbf{p} \gets \mathbf{N}[0]$;
			$\mathbf{N} \gets \mathbf{N} - \mathbf{p}$;
			$\mathbf{O} \gets \mathbf{O} - \mathbf{p}$;
			$\mathbf{c} \gets \mathbf{c} + \{\mathbf{p}\}$\;
			\ForEach{$\mathbf{p}^{'} = Neighbor(\mathbf{p})  \in \mathbf{O}$ \emph{and} $Distance(\mathbf{p}^{'}, \mathbf{p}) \textless T_d$}
			{
				$\mathbf{N} \gets \mathbf{N} + Neighbor(\mathbf{p}^{'})$ \;	
			}
		}
		$N_\mathit{pixel} \gets $ the number of pixels in $\mathbf{c}$\;
		\If{$N_\mathit{pixel} \textless T_n$}
		{
		    $\mathbf{C} \gets \mathbf{C} - \mathbf{c}$\;
		}
	}
	\ForEach{$\mathbf{c} \in \mathbf{C}$}
	{
		$w, h \gets Width(\mathbf{c}), Height(\mathbf{c})$\;
		$N_\mathit{SmallR} \gets $ the number of pixels in $\mathbf{c}$ whose range value is smaller than its neighbor outside $\mathbf{c}$ \;
		\If{$h / w \textless 1$ \emph{or} $N_\mathit{SmallR} \textless \delta \cdot Len(\mathbf{c}) $}
		{
			$\mathbf{C} \gets \mathbf{C} - \mathbf{c}$\;
		}
	}
	\ForEach{$\mathbf{c} \in \mathbf{C}$}
	{
		$\mathbf{x}, \mathbf{y}, \mathbf{z} \gets $ 3D coordinates of pixels in $\mathbf{c}$\;
		\If{$\max(\mathbf{z}) \textgreater H_a$ \emph{and} $\min(\mathbf{z}) \textless H_b$ \emph{and} $(\max(\mathbf{z})-\min(\mathbf{z})) \textgreater T_h$}
		{
			$x_\mathbf{c}, y_\mathbf{c}, r_\mathbf{c} \gets FitCircle(\mathbf{x},\mathbf{y})$ \;
			$N_\mathit{Free} \gets $ the number of the pixels in a small free space outside the radius of the pole\;
			\If{$r_\mathbf{c} \textless R_a$ \emph{and} $r_\mathbf{c} \textgreater R_b$ \emph{and} $N_\mathit{Free} \textless \mu \cdot Len(\mathbf{z})$ } 
			{
				$\mathbf{P} \gets \mathbf{P} + \{ x_\mathbf{c}, y_\mathbf{c}, r_\mathbf{c} \}$
			}
		}
	}
  \caption{Range Image-based Pole Extraction}
  \label{alg:pole_extractor}
\end{algorithm2e}

\begin{figure}[t]
	\centering
	\begin{subfigure}[b]{\linewidth}
	  \includegraphics[width=\linewidth]{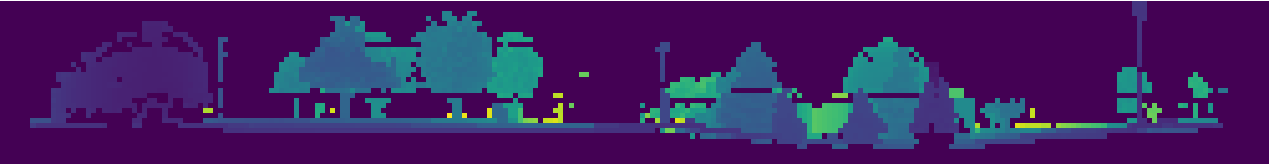}
	\end{subfigure}
	\begin{subfigure}[b]{\linewidth}
	  \includegraphics[width=\linewidth]{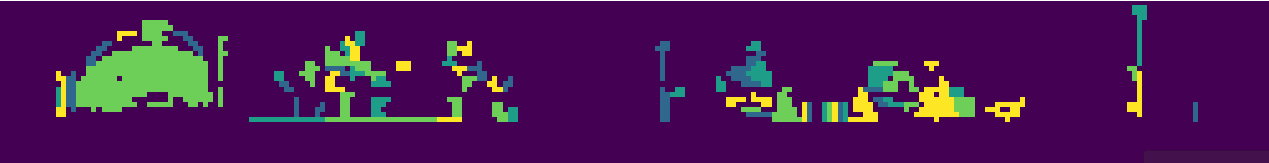}
	\end{subfigure}
	\begin{subfigure}[b]{\linewidth}
		\includegraphics[width=\linewidth]{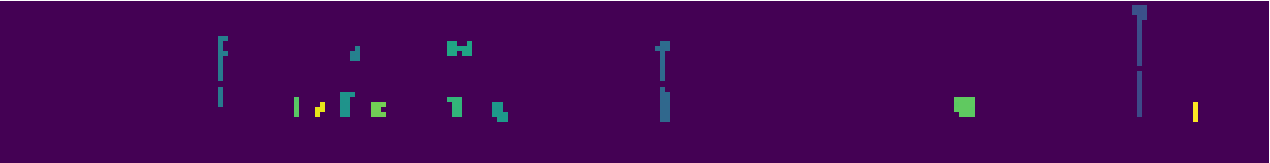}
	  \end{subfigure}
	\begin{subfigure}[b]{\linewidth}
		\includegraphics[width=\linewidth]{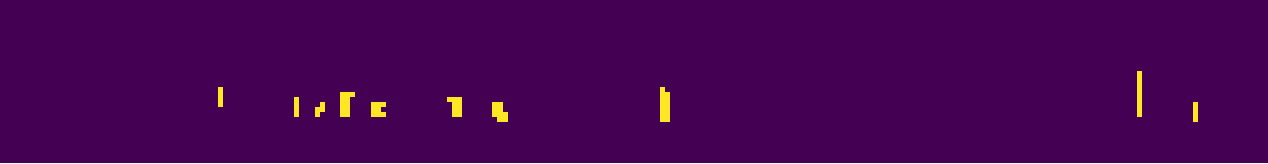}
	  \end{subfigure}
	\caption{Visualization of results on each step of our geometric pole extractor. The first image shows the range image. The second image represents the clustering result and the third one shows the pole candidates after applying 2D geometric constraints. The last one is the final pole extraction result.}
	\label{fig:pole_extractor_vis}
  \end{figure}

We extract poles based on the range images generated in the previous step. 
The general intuition behind our pole extraction algorithm is that the range values of the poles are usually significantly smaller than the backgrounds. 
Based on this idea and as specified in~\algref{alg:pole_extractor}, our first step is to cluster the pixels of the range image into different small regions based on their range values. 
We first pass through all pixels in the range image, from top to bottom, left to right. 
We put all pixels with valid range data in an open set~$\mathbf{O}$. 
For each valid pixel~$\mathbf{p}$, we check its neighbors including the left, right and below ones.
If there exists a neighbor with a valid value and the range difference between the current pixel and its neighbor is smaller than a threshold~$T_d$, we add the current pixel to a cluster set~$\mathbf{c}$ and remove it from the open set~$\mathbf{O}$. 
We do the same check iteratively with the neighbors until no neighbor pixel meets the above criteria, and we then get a cluster of pixels.
After checking all the pixels in~$\mathbf{O}$, we will get a set~$\mathbf{C}$ with several clusters and each cluster represents one object.
If the number of pixels in one cluster is smaller than a threshold $T_n$, we regard it as an outlier and ignore it.

The next step is to extract poles from these objects using 2D geometric constraints.
To this end, we exploit both the range information and the 3D coordinates~$(x, y, z)$ of each pixel.
We first check the aspect ratio of each cluster.
Since we are only interested in pole-like objects, whose height is usually larger than its width, we therefore discard the cluster with aspect ratio~$h / w \textless 1$.
Another heuristic we use is the fact that a pole usually stands alone and has a significant distance from background objects.
$N_\mathit{SmallR}$ is the number of points in cluster~$\mathbf{c}$ whose range value is smaller than its neighbor outside~$\mathbf{c}$,
we discard the cluster if $N_\mathit{SmallR}$ is smaller than $\delta$ times the number of all points in the cluster.

To exploit the 3D coordinates~$(x, y, z)$ of each pixel, 
we calculate $\max(z) - \min(z)$ of each cluster and only take a cluster as a pole candidate if $\max(z) - \min(z) > T_h$.
Besides, we are only interested in poles whose height is higher than~$H_a$.
Based on experience, we also set a threshold~$H_b$ for the lowest position of the pole to filter outliers.
For each pole candidate, we then fit a circle using the~$x$ and~$y$ coordinates of all points in the cluster and get the center and the radius of that pole.
We filter out the candidates with too small or too large radiuses and candidates that connect to other objects by checking the free space around them.
After the above steps, we finally extract the positions and radiuses of poles. As an example, \figref{fig:pole_extractor_vis} visualize the intermediate results on each step of our geometric pole extractor.

\subsection{Learning-based Pole Segmentation Trained With Pseudo Labels}
\label{poleExLearning}

\begin{figure*}[htb]
\begin{center}
\includegraphics[width=0.9\linewidth]{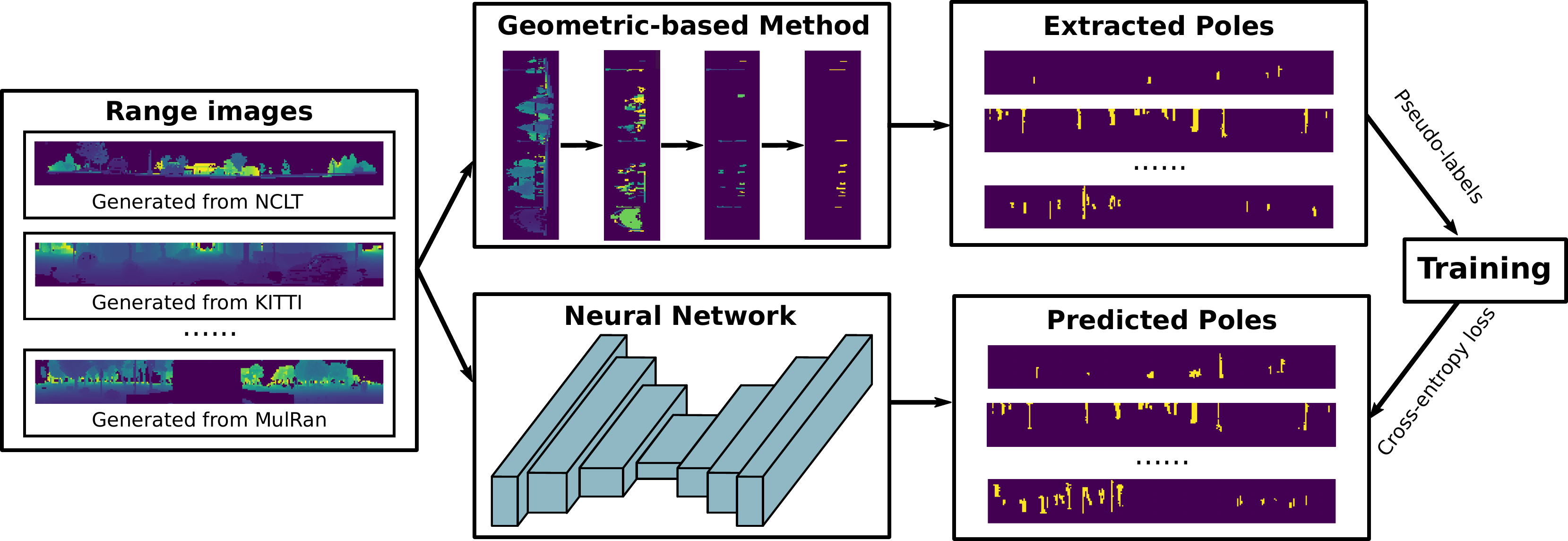}
\end{center}
\caption{Overview of our learning-based pole extraction approach. We first use the geometric method to generate range images and pseudo pole labels on multiple datasets, including NCLT, KITTI, and MulRan. Then, we train a deep neural network to extract pole-like objects directly on the range images. Our learning-based pole extractor generalizes well to all three datasets with a single model.}
\label{fig:pole_extractor_l}
\end{figure*}

As shown in~\cite{chen2022ral}, geometric information can be used to automatically generate labels for training a LiDAR-based moving object segmentation network and achieve good performance in various environments. 
Such auto-labeling methods enable network learning in a self-supervised manner, which saves the extensive manual labeling effort and improves the generalization ability of the learning-based method.
Inspired by it, we use the poles detected by our geometry-based pole extractor to generate pseudo labels to train an online pole segmentation network.

We use our geometry-based method to generate pseudo pole labels from NCLT~\cite{carlevaris-bianco2016ijrr}, SemanticKITTI~\cite{behley2019iccv}, and MulRan~\cite{kim2020icra} datasets. 
In this work, we do not design a new network architecture but reuse networks that have been successfully applied to LiDAR-based semantic segmentation in the past. We adopt and evaluate SalsaNext~\cite{cortinhal2020iv}, an encoder-decoder architecture with solid performances on semantic segmentation tasks. SalsaNext~\cite{cortinhal2020iv} achieves state-of-the-art performances on SemanticKITTI dataset among all range image-based semantic segmentation networks. Therefore, we choose it as the base network architecture for our learning-based method. 
Instead of segmenting the environment into multiple classes like ground, structure, vehicle and human, in our case, we only distinguish the poles from other objects. After the segmentation, similar filtering steps as used in the geometric method are applied to remove outliers. SalsaNext network is comparably light-weight, and can achieve real-time operation, i.e., run faster than the commonly used frame rate of the employed LiDAR sensor, which is $10\,\mathrm{Hz}$ for Ouster and Velodyne scanners. For more detailed information about the network, we refer to the original paper~\cite{cortinhal2020iv}. 

For training the segmentation network, we directly feed them with the range images plus the pseudo pole labels generated from our geometry-based pole extractor.
We use the same loss functions as used in the original segmentation methods, while mapping all classes into two per-point classes, poles and non-poles.
We retrain the network and evaluate the pole extracting performance with our pole datasets and also localization tasks.
\figref{fig:pole_extractor_l} shows the training pipeline of our proposed learning-based pole segmentation method. 
Note that, we train the network with pseudo pole labels generated from different datasets, and later use the same model to extract poles in different environments.

\subsection{Pole-based Mapping\label{map}}
To build the global map for localization, we follow the same setup as introduced by  Schaefer \etal~\cite{schaefer2019ecmr}, splitting the ground-truth trajectory into shorter sections with equal length, extracting poles in these sections separately and finally merging them into a global pole map.
Since the provided poses are not very accurate for mapping~\cite{schaefer2019ecmr}, instead of aggregating a noisy submap, we only use the middle LiDAR scan of each section to extract poles.
We merge multiple overlapped pole detections by averaging over their centers and radiuses and apply a counting model to filter out the dynamic objects. Only those candidate poles that appear multiple times in continuous sections are added to the map.

\subsection{Monte Carlo Localization\label{loc}}

Monte Carlo localization~(MCL) is commonly implemented using a particle filter~\cite{dellaert1999icra}. 
MCL realizes a recursive Bayesian filter estimating a probability density~$p(\v{x}_t\mid\v{z}_{1:t},\v{u}_{1:t})$ over the pose~$\v{x_t}$ given all observations~$\v{z}_{1:t}$ and motion controls~$\v{u}_{1:t}$ up to time~$t$. This posterior is updated as follows:
\begin{align}
  &p(\v{x}_t\mid\v{z}_{1:t},\v{u}_{1:t}) = \eta~p(\v{z}_t\mid\v{x}_{t}) \cdot\nonumber\\
  &\;\;\int{p(\v{x}_t\mid\v{u}_{t}, \v{x}_{t-1})~p(\v{x}_{t-1} \mid \v{z}_{1:t-1},\v{u}_{1:t-1})\ d\v{x}_{t-1}},
\label{eq:bayesian}
\end{align}
where~$\eta$ is a normalization constant, $p(\v{x}_t\mid\v{u}_{t}, \v{x}_{t-1})$~is the motion model, $p(\v{z}_t\mid\v{x}_{t})$~is the observation model, and $p(\v{x}_{t-1} \mid \v{z}_{1:t-1},\v{u}_{1:t-1})$~is the probability distribution for the prior state $\v{x}_{t-1}$.

In our case, each particle represents a hypothesis for the 2D pose $\v{x}_t = (x, y, \theta)_t$ of the robot at time~$t$. 
When the robot moves, the pose of each particle is updated based on a motion model with the control input~$\v{u}_t$ or the odometry measurements.
For the observation model, the weights of the particles are updated based on the difference between expected observations and actual observations.
The observations are the positions of the poles. We match the online observed poles with the poles in the map via nearest-neighbor search using a k-d tree. 
The likelihood of the $j$-th particle is then approximated using a Gaussian distribution:
\begin{align}
  p \left(\v{z}_{t} \mid \v{x}_{t} \right) &\propto \prod_{i}^{N} \left(\exp{\left(-\frac{1}{2} \frac{{d\left(\v{z}^{i}_{t}, \v{z}^{i}_{j} \right)}^2}{\sigma^2_d}\right)+\epsilon}\right),
\label{eq:sensor}
\end{align}
where~$N$ is the number of matched poles in the current scan, $\sigma^2_d$ is the position uncertainty of the poles, $d$ corresponds to the difference between the online observed pole~$\v{z}^{i}_{t}$ and matched pole in the map $\v{z}^{i}_{j}$ given the position of the  particle~$j$.
We use the Euclidean distance between the pole positions to measure this difference. The constant $\epsilon$ accounts for the probability that a detected pole is not part of the map. This constant is crucial for the robustness of localization when there are many outliers.
If the number of effective particles decreases below a threshold~\cite{grisetti2007tro}, the resampling process is triggered and particles are sampled based on their weights.

\section{Experimental Evaluation}
\label{sec:exp}

The main focus of this work is an accurate and efficient pole extractor for long-term LiDAR localization.
We present our experiments to show the capabilities of our method. The experiments furthermore support our key claims that our method is able to:
(i)~extract more reliable poles in the environment compared to the baseline method, as a result, 
(ii)~achieve better online localization performance in different environments, and 
(iii)~generate pseudo pole labels to train a pole segmentation network achieving better localization results and faster runtime compared to the geometric method.

\subsection{Datasets for Pole Extraction and LiDAR Localization}

There are few public datasets available to evaluate pole extraction performance. 
To this end, we label the poles in session 2012-01-08 of NCLT dataset by hand and release this dataset for public research use. 
For the reason that the original NCLT ground-truth poses are inaccurate~\cite{schaefer2019ecmr}, the aggregated point cloud is a little blurry. Therefore, to create the ground-truth pole map of the environment, we partition the ground-truth trajectory into shorter segments of equal length. 
For each segment, we aggregate the point cloud together and use Open3D~\cite{zhou2018arXiv} to render and label the pole positions. 
We only label those poles with high certainty and ignore those blurry ones.
Besides our own labelled data, we also reorganize the SemanticKITTI~\cite{behley2019iccv} dataset sequences 00-10 by extracting the pole-like objects like traffic signs, poles and trunk, and then clustering the point clouds to generate the ground-truth pole instances. 

\begin{figure*}[t]
	\centering
	\begin{subfigure}[b]{0.245\linewidth}
	  \includegraphics[width=\linewidth]{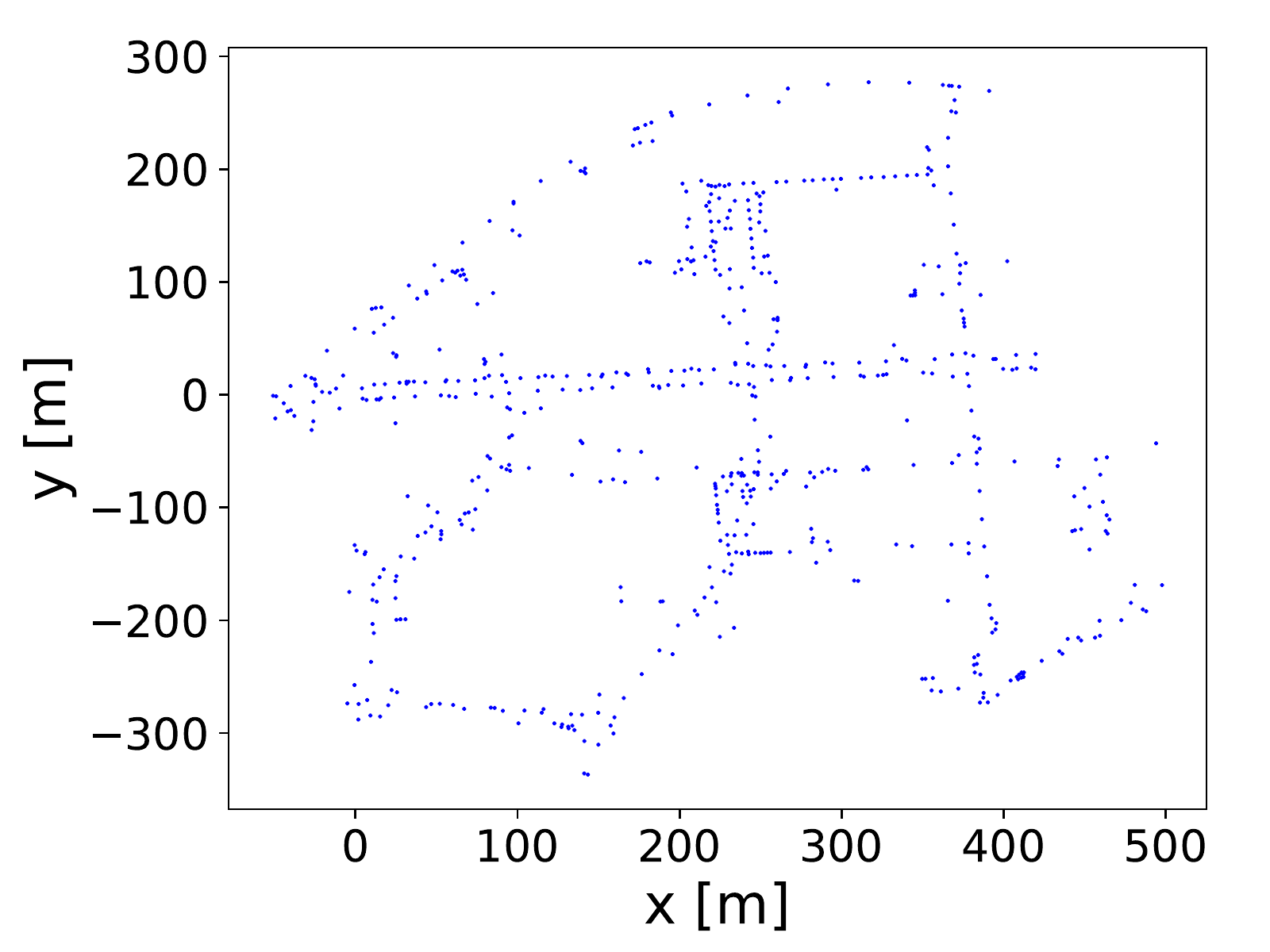}
	\end{subfigure}
	\begin{subfigure}[b]{0.245\linewidth}
	  \includegraphics[width=\linewidth]{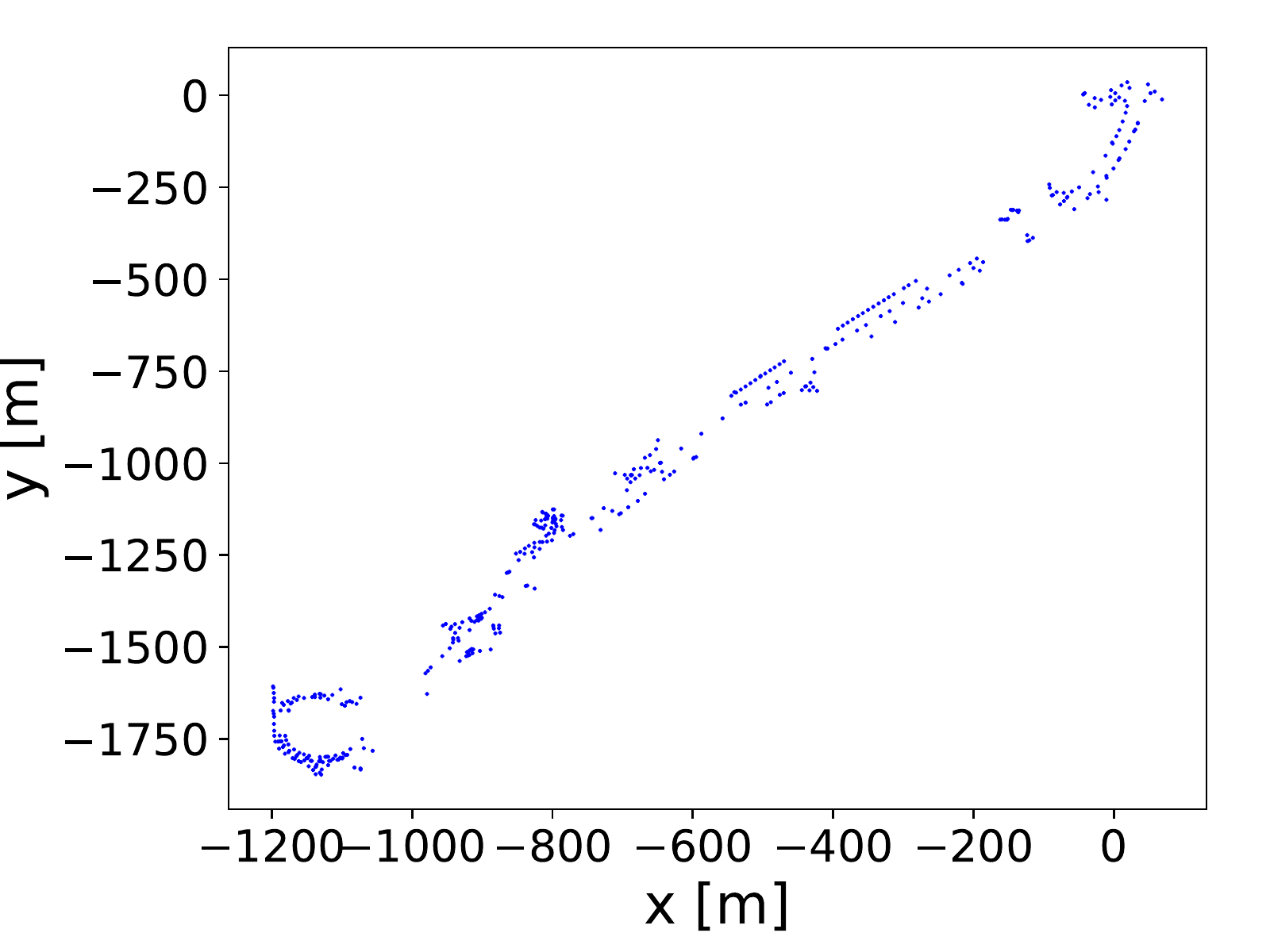}
	\end{subfigure}
	\begin{subfigure}[b]{0.245\linewidth}
		\includegraphics[width=\linewidth]{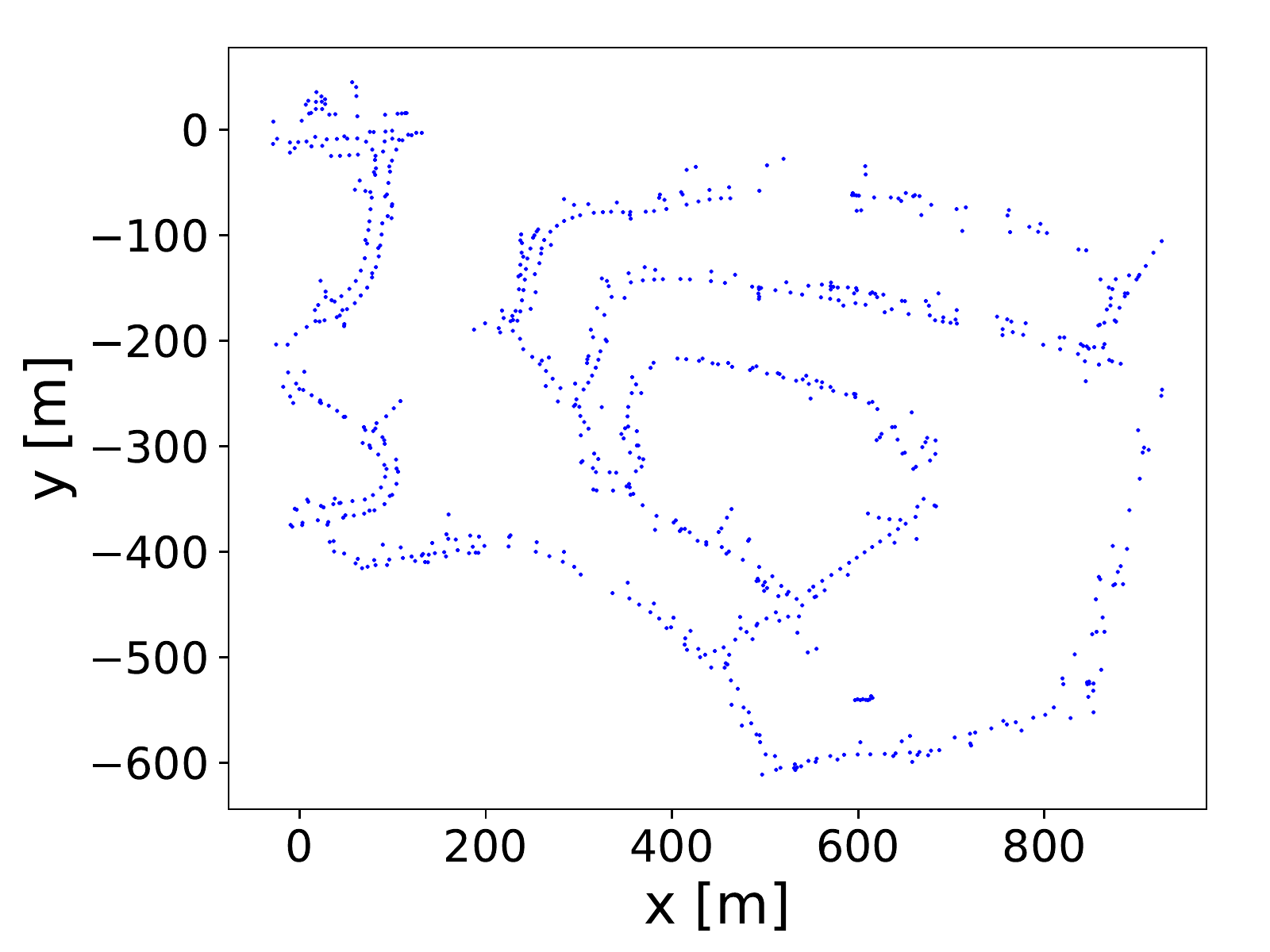}
	  \end{subfigure}
	\begin{subfigure}[b]{0.245\linewidth}
		\includegraphics[width=\linewidth]{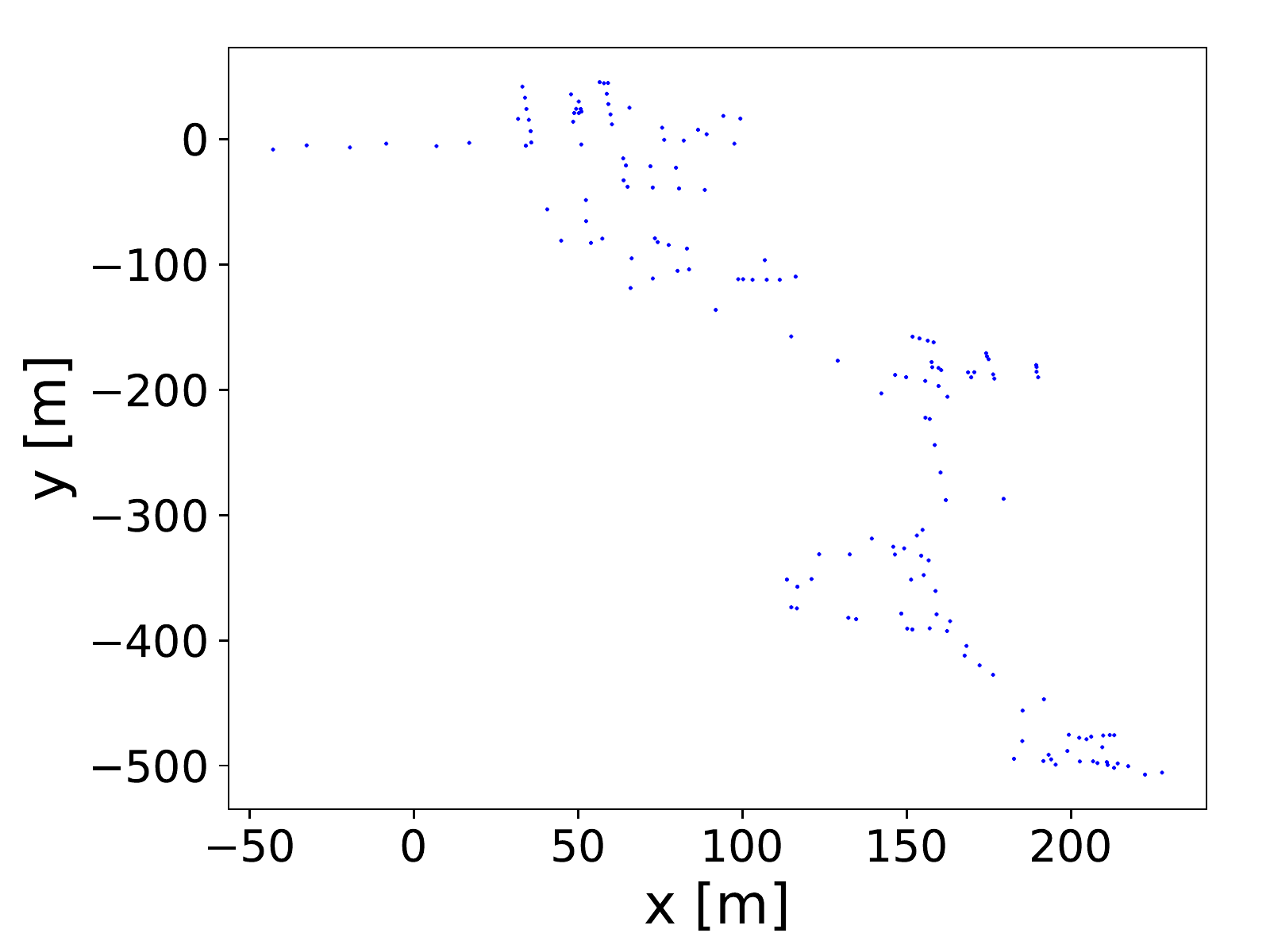}
	  \end{subfigure}
	\begin{subfigure}[b]{0.245\linewidth}
		\includegraphics[width=\linewidth]{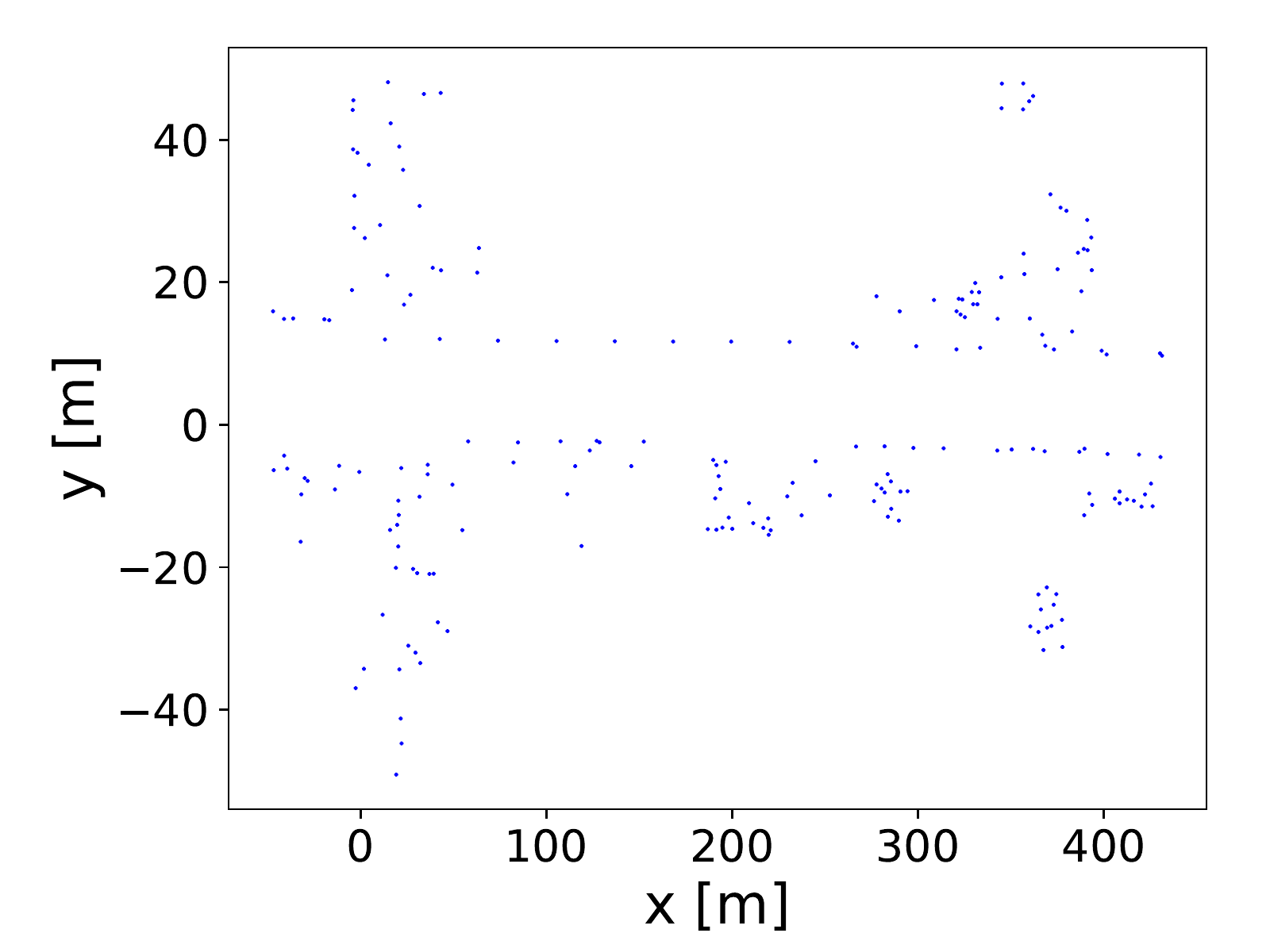}
	  \end{subfigure}
	\begin{subfigure}[b]{0.245\linewidth}
		\includegraphics[width=\linewidth]{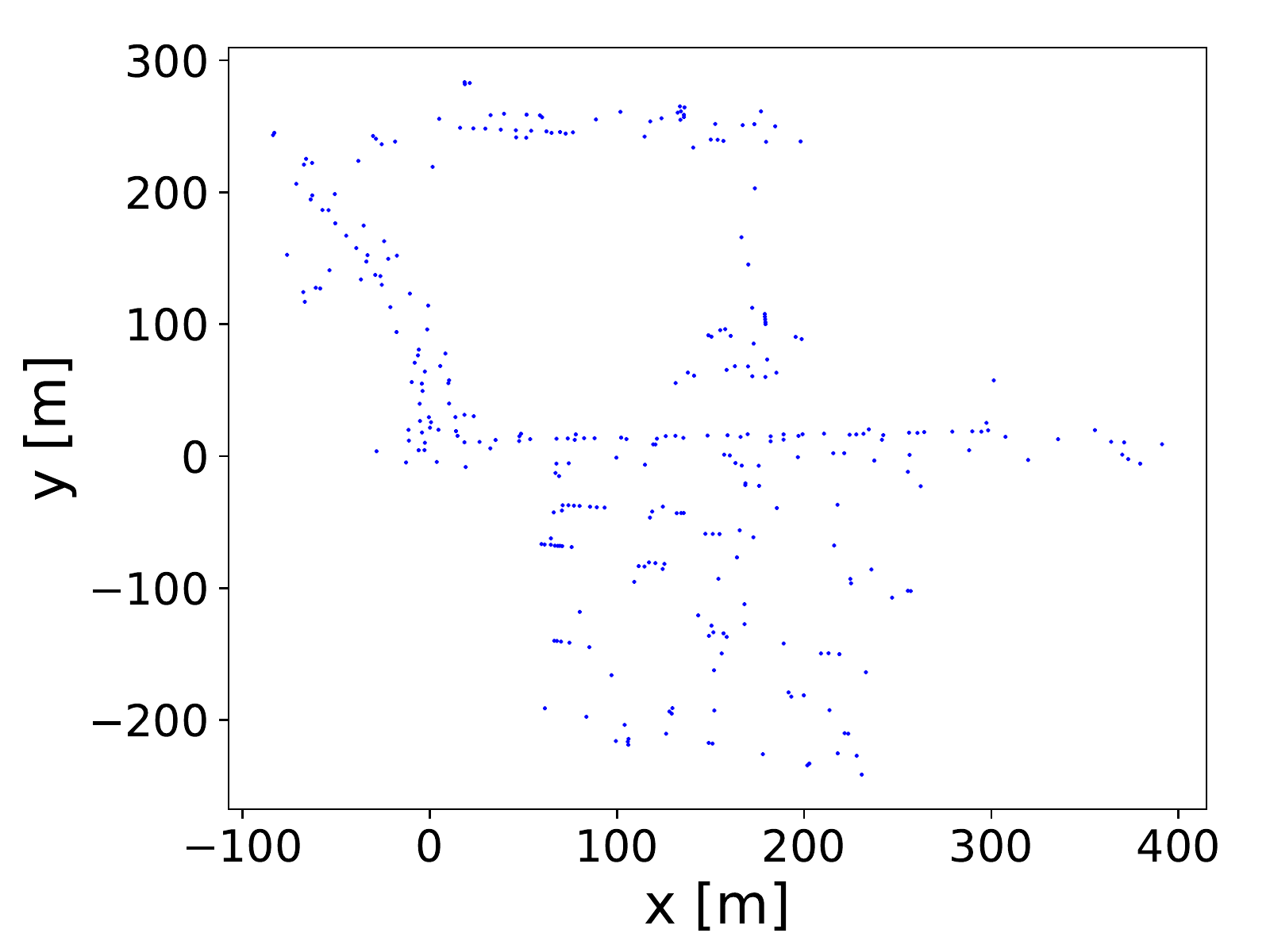}
	  \end{subfigure}
	\begin{subfigure}[b]{0.245\linewidth}
		\includegraphics[width=\linewidth]{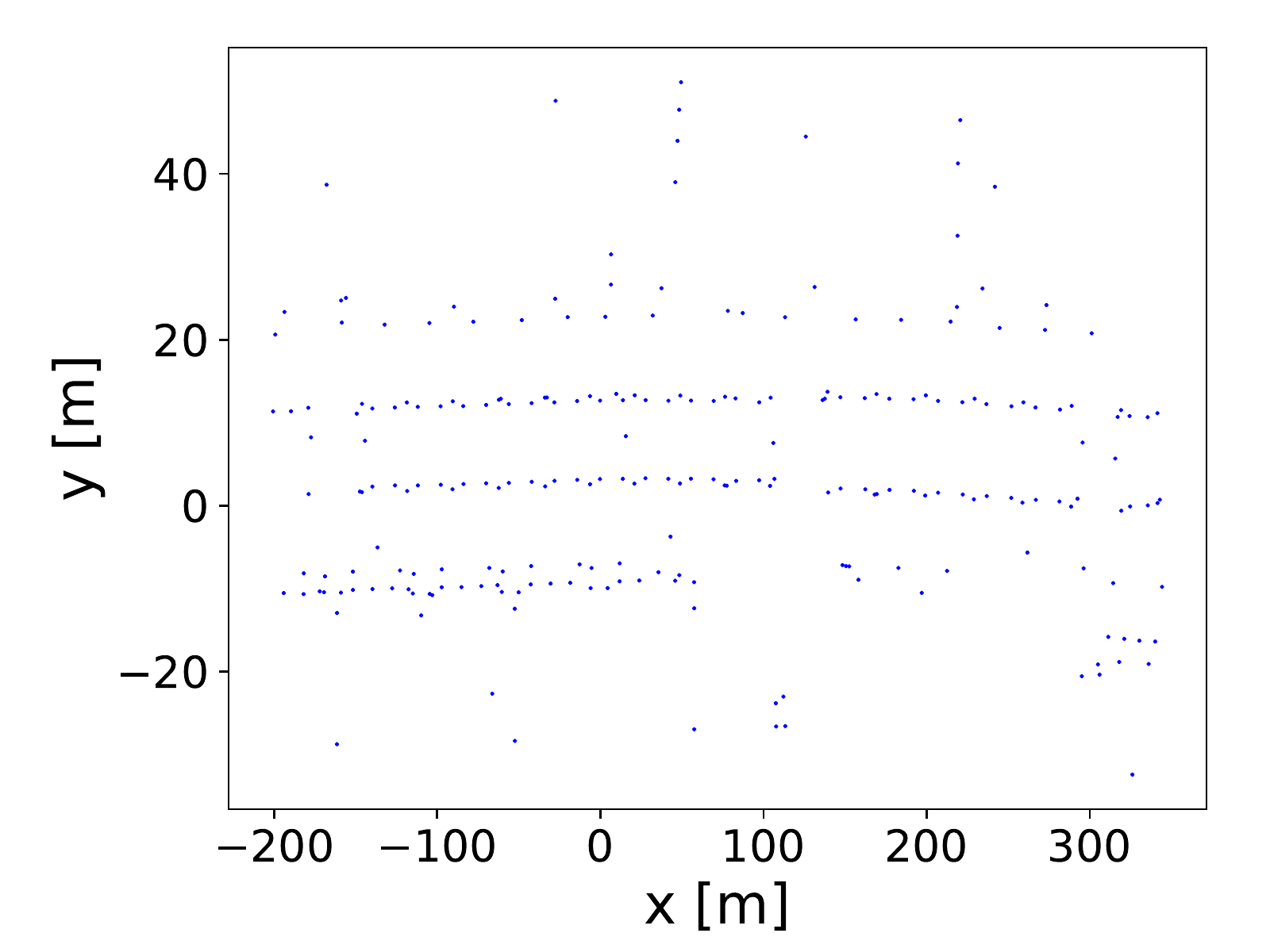}
	  \end{subfigure}
	\begin{subfigure}[b]{0.245\linewidth}
		\includegraphics[width=\linewidth]{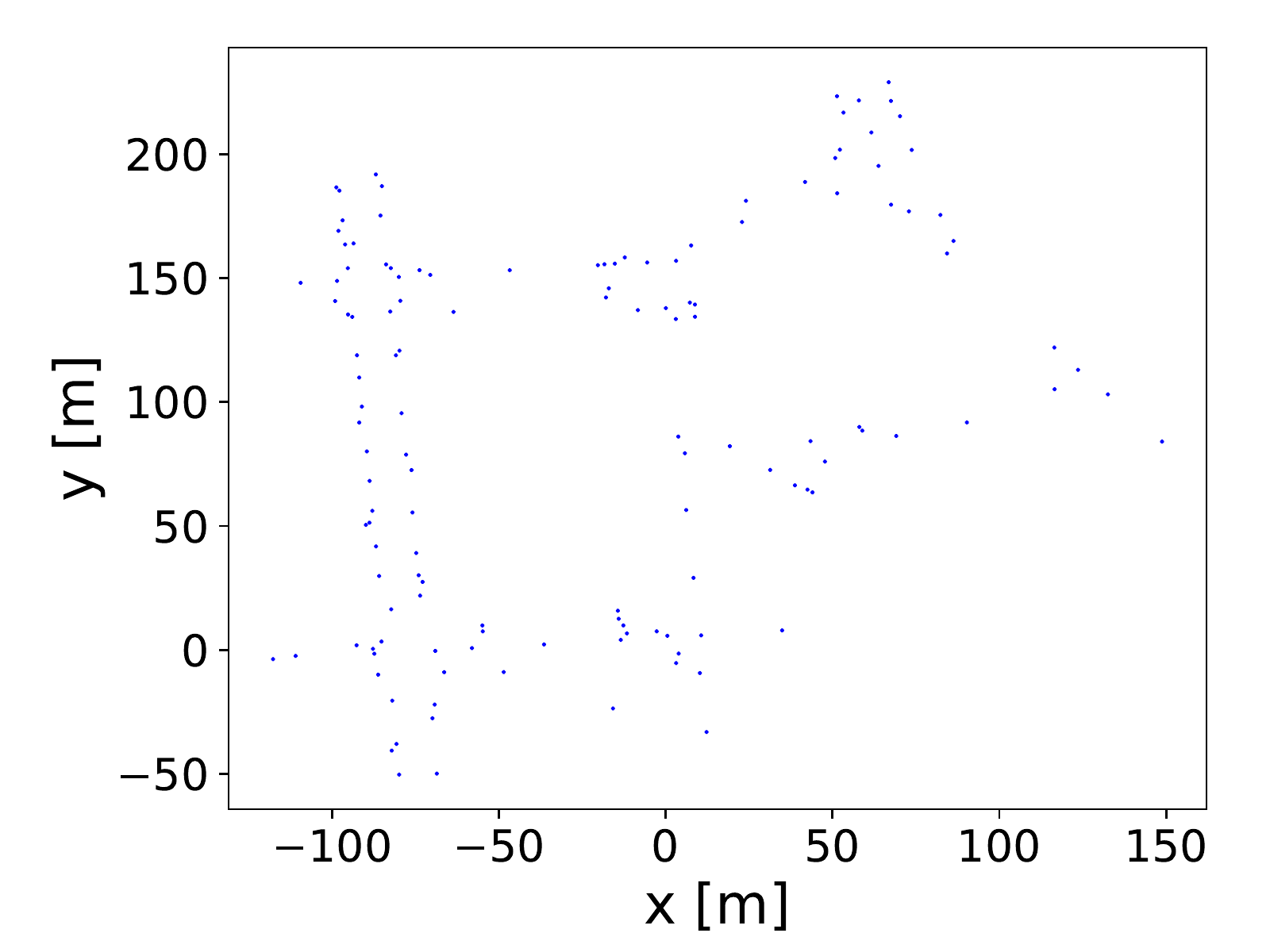}
	  \end{subfigure}
	\begin{subfigure}[b]{0.245\linewidth}
		\includegraphics[width=\linewidth]{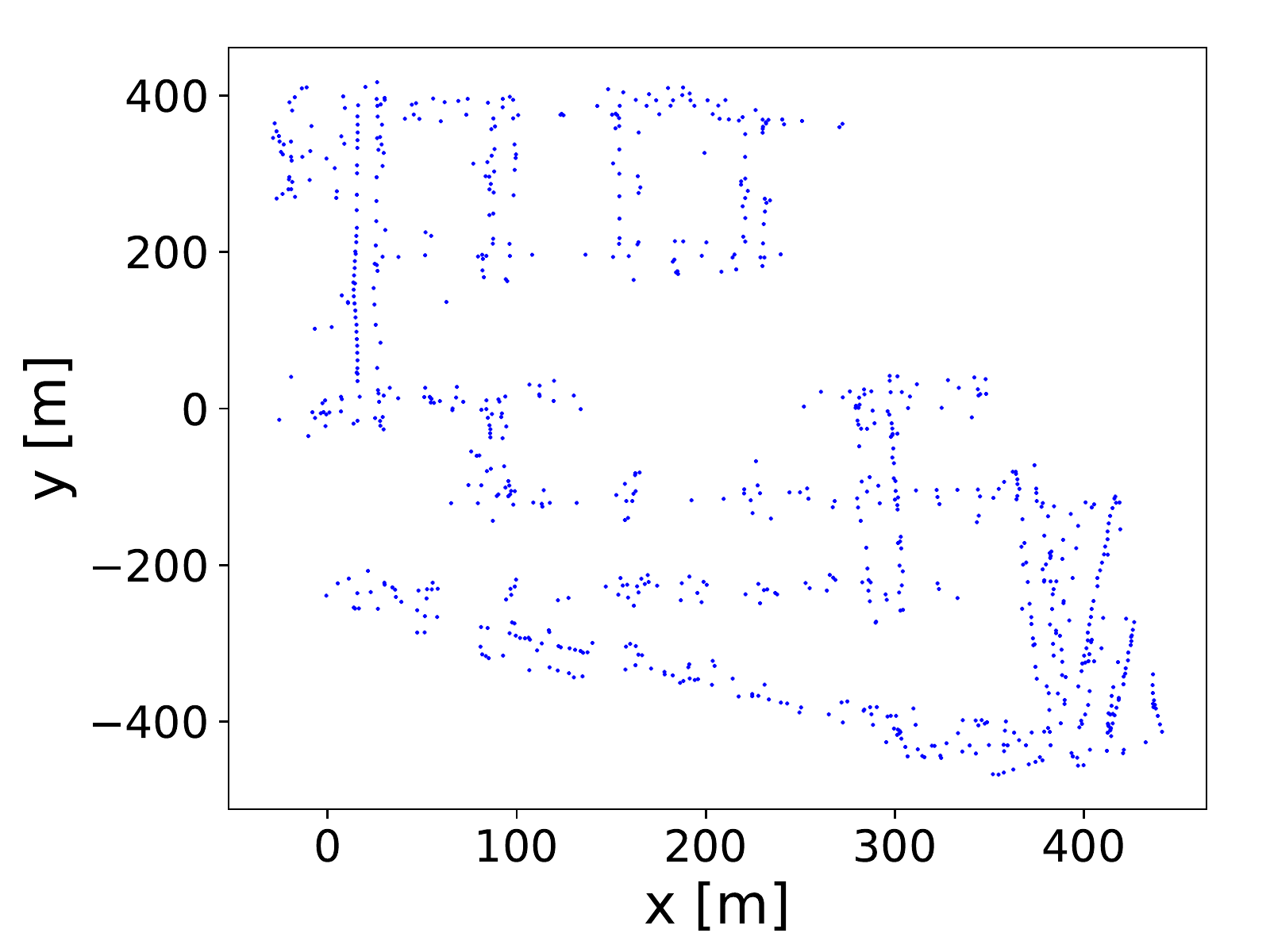}
	  \end{subfigure}
	\begin{subfigure}[b]{0.245\linewidth}
		\includegraphics[width=\linewidth]{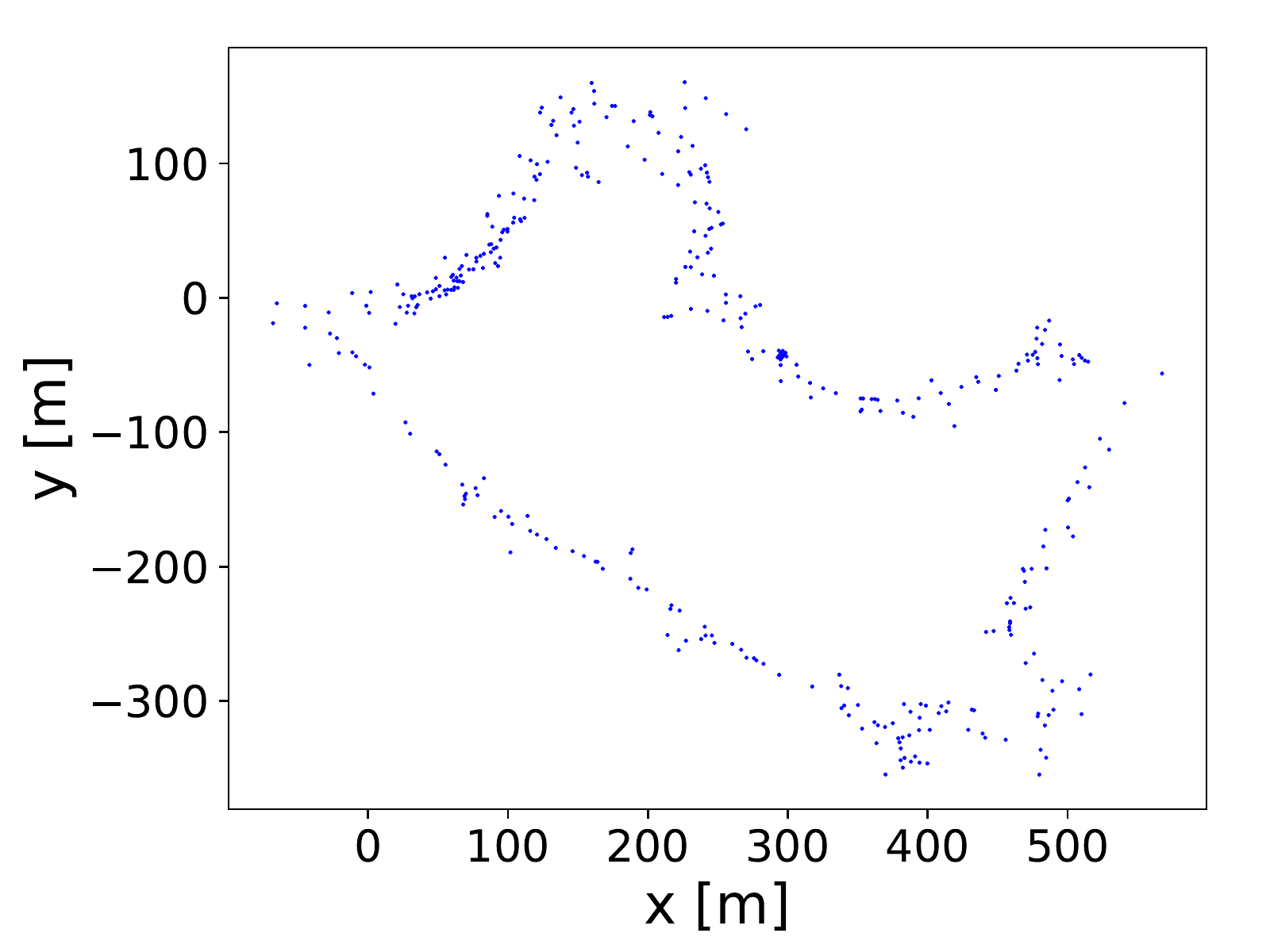}
	  \end{subfigure}
	\begin{subfigure}[b]{0.245\linewidth}
		\includegraphics[width=\linewidth]{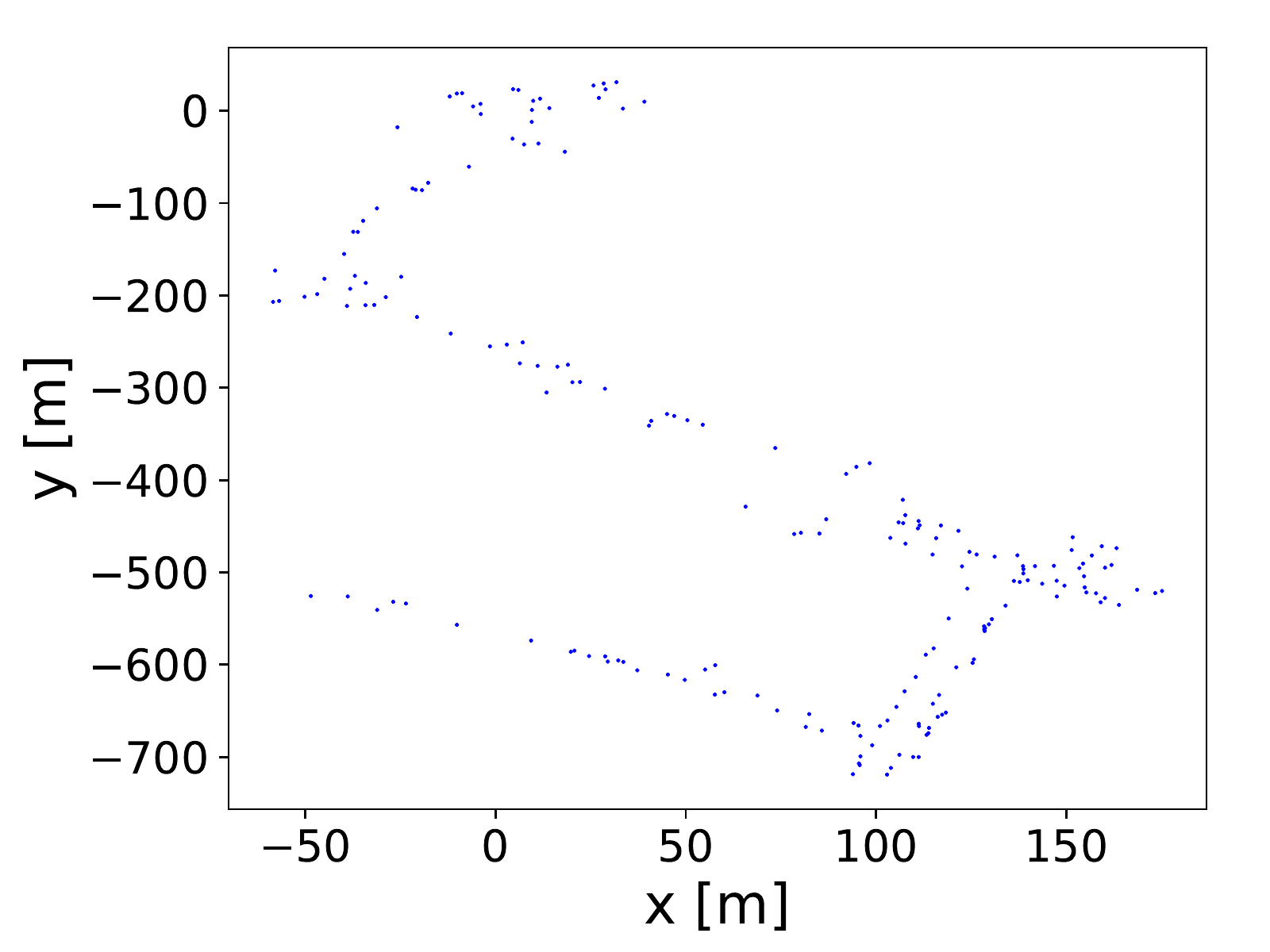}
	  \end{subfigure}
	\begin{subfigure}[b]{0.245\linewidth}
		\includegraphics[width=\linewidth]{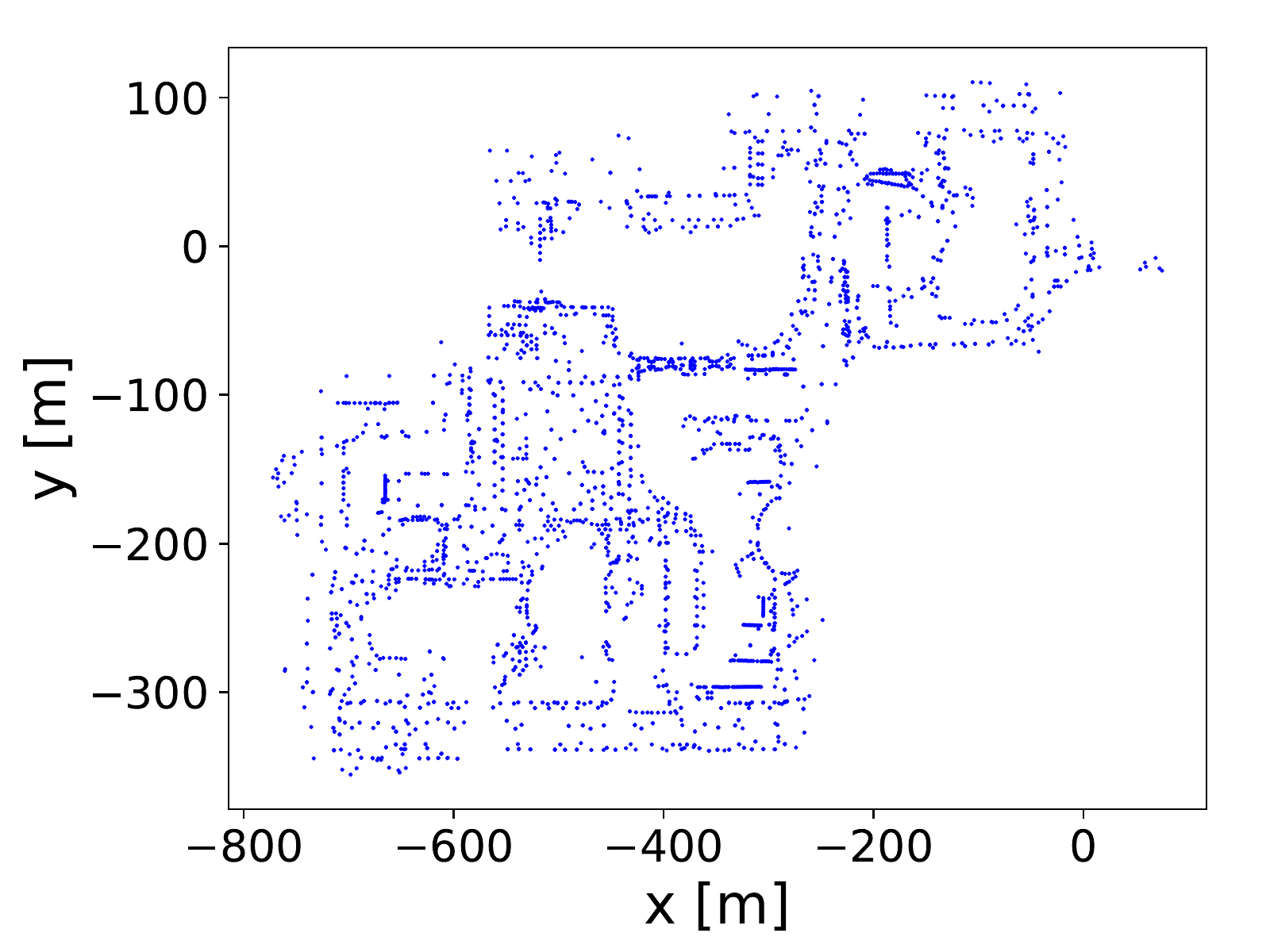}
	  \end{subfigure}
	\caption{Our pole extraction datasets. The first 11 figures show the ground truth pole positions of sequence 00-10 of SemanticKITTI, and the last one (at the right bottom corner) shows pole positions of the session 2012-01-08 of NCLT. Each blue dot represents the position of one pole.}
	\label{fig:pole_result_vis}
  \end{figure*}

To assess the localization reliability and accuracy of our method, we use NCLT dataset~\cite{carlevaris-bianco2016ijrr} and MulRan dataset~\cite{kim2020icra}. 
These two datasets are collected in different environments (U.S., Korea) with different LiDAR sensors (Velodyne HDL-32E, Ouster OS1-64).
In these two datasets, the robot passes through the same place multiple times with month-level temporal gaps, hence ideal to test the long-term localization performance. 
We compare our methods to both a pole-based method proposed by Schaefer \etal~\cite{schaefer2019ecmr} and the range image-based method proposed by Chen \etal~\cite{chen2021icra}. We reproduce their results using the public available codes.
For the SemanticKITTI dataset, there is no overlap area between different sequences for evaluating long-term localization.
Therefore, we only used the extracted pole labels from the SemanticKITTI dataset to train our network. 
\figref{fig:pole_result_vis} shows examples of our proposed pole datasets.

\subsection{Pole Extractor Performance}

The first experiment evaluates the pole extraction performance of our approach and supports the claim that our range image-based method outperforms the baseline method in pole extraction. 

We evaluate both our geometry-based pole extractor, named Ours-G, and our learning-based pole segmentation method, named Ours-L.
For training the pole segmentation network, we use data from multiple datasets, including the session 2012-01-08 in NCLT dataset, sequence KAIST 02 in MulRan dataset and sequence 00-02, 05-09 in SemanticKITTI dataset. For validation, we use sequences 03 and 04 in SemanticKITTI dataset and sequence 10 for testing. We train the network for 150 epochs using stochastic gradient descent with an initial learning rate of 0.01 and the learning rate decay is 0.01. The batch size is 12 and the spatial dropout probability is 0.2. The size of the range image is 32$\times$256 and the valid range values are normalized between 0 and 1. To prevent overfitting, we augmented the data by applying a random rotation or translation, flipping randomly around the y-axis with a probability of 0.5.
During the matching phase, we find the matches via nearest-neighbor search using a k-d tree with $1\,\mathrm{m}$ distance bounds.

\begin{figure*}[htb]
\begin{center}
\includegraphics[width=\linewidth]{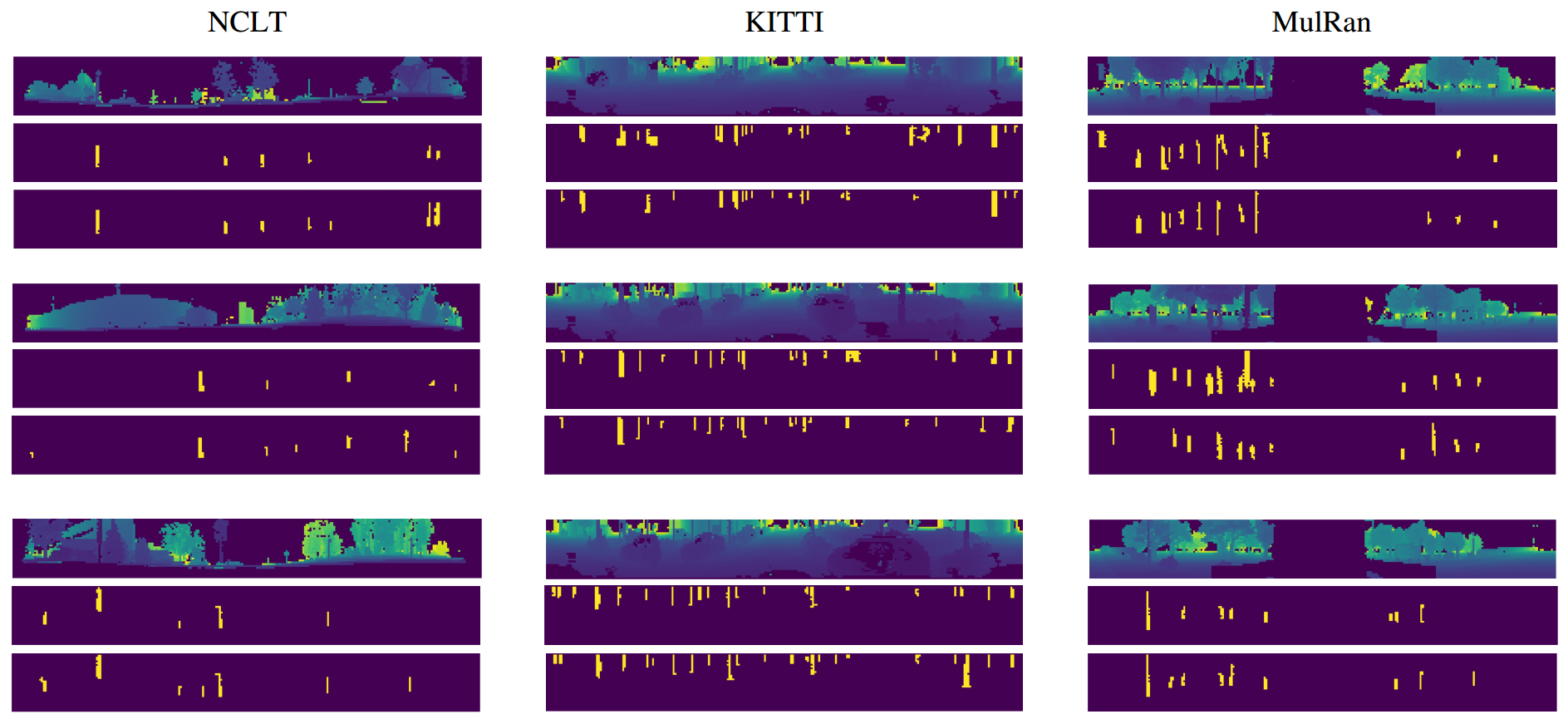}
\end{center}
\vspace{-2mm}
\caption{Pole extraction examples using our geometric and learning-based pole extractor. For each sample, the first figure is the range image, the second one is the pole extraction results by our geometric method, and the last one shows the pole extraction results by our learning-based method. Note that, for the learning-based method we get all the results on three datasets with a single trained model.}
\label{fig:pole_extractor_l_samples}
\vspace{-3mm}
\end{figure*}

\tabref{tab:pole_extractor} summarizes the precision, recall and F1 score of our method and Schaefer \etal~\cite{schaefer2019ecmr} with respect to the ground-truth pole map on both NCLT dataset and SemanticKITTI dataset.
As can be seen, our methods achieve better performance and extract more poles in both environments compared to the baseline method.
Compared to our geometry-based pole extractor, our learning-based method finds more poles while introducing more false positives, which decreases precision.
This can also be seen in~\figref{fig:pole_extractor_l_samples}, which shows pole extraction examples of our geometric and learning-based pole extractor.

Note that we trained our pole segmentation network only once with pseudo pole labels generated from different datasets and evaluated it on multiple different datasets. 
As can be seen in~\figref{fig:pole_extractor_l_samples}, the environments of different datasets vary a lot, while our learning-based method can still extract poles well without fine-tuning, which shows a good generalization ability of our method. 
The possible reason for that is the range values of the poles are usually significantly different than the backgrounds, which makes poles distinctive and easy to be detected on range images. 
Compared to multi-class segmentation, it is easier for the neural network to learn a more general model to detect poles based on the range images~\cite{chen2021ral}.
Furthermore, the learning-based method has a higher recall than the geometry-based method, but with a lower precision, which means that the learning-based method detects more true positives, but also more false positives. We use the detected poles as landmarks for MCL, which is a very robust probabilistic localization system. Thus, the localization performance will not be influenced by little false positives, but benefits from higher recalls with more landmarks, as shown in the next section.

\begin{table}[t]
\caption{Pole Extraction Precision, Recall, and F1 Score on NCLT and KITTI datasets.}
\centering
\scalebox{1}{
\scriptsize{
\begin{tabular}{ll|ccc}
	\toprule
    Dataset & Method & Precision & Recall & F1 Score \\
	\midrule
	\multirow{3}{1cm}{NCLT} & Schaefer~\cite{schaefer2019ecmr} & 0.690  & 0.386 & 0.495 \\
	& Ours-G & $\mathbf{0.765}$  & 0.657 &  $\mathbf{0.706}$\\
	& Ours-L & 0.675  & $\mathbf{0.674}$ &  0.674\\
	\midrule
	\multirow{3}{1cm}{Semantic KITTI}  & Schaefer~\cite{schaefer2019ecmr} &0.621 &0.380 & 0.455\\
	& Ours-G & $\mathbf{0.687}$ & 0.439 &0.515 \\
	& Ours-L &0.607 & $\mathbf{0.582}$ & $\mathbf{0.594}$ \\
	\bottomrule
\end{tabular}
}
}
\label{tab:pole_extractor}
\end{table}

\subsection{Localization Performance}
The second experiment is presented to support the claim that our approach achieves higher accuracy on localization in different environments.
For all the experiments, we use the same setup as used in the baselines and report their results from the original work.

\subsubsection{Localization on the NCLT Dataset}

\begin{table*}
	\centering
	\small
	\scalebox{0.8}{
	\setlength{\tabcolsep}{3pt}
	\begin{tabular}{l |
			S[round-mode=places, round-precision=3]
			S[round-mode=places, round-precision=3]
			S[round-mode=places, round-precision=3] |
			S[round-mode=places, round-precision=3]
			S[round-mode=places, round-precision=3]
			S[round-mode=places, round-precision=3] |
			S[round-mode=places, round-precision=3]
			S[round-mode=places, round-precision=3]
			S[round-mode=places, round-precision=3] |
			S[round-mode=places, round-precision=3]
			S[round-mode=places, round-precision=3]
			S[round-mode=places, round-precision=3]}
		\toprule
		Session Date 
			& \multicolumn{3}{c}{$\Delta_{\mathrm{pos}}$}
			& \multicolumn{3}{c}{$\RMSE_{\mathrm{pos}}$}
			& \multicolumn{3}{c}{$\Delta_{\mathrm{ang}}$}
			& \multicolumn{3}{c}{$\RMSE_{\mathrm{ang}}$} \\
			& \multicolumn{3}{c}{[\si{m}]}
			& \multicolumn{3}{c}{[\si{m}]}
			& \multicolumn{3}{c}{[\si{\degree}]}
			& \multicolumn{3}{c}{[\si{\degree}]} \\
		\toprule
			& \text{Schaefer~\cite{schaefer2019ecmr}} & \text{Ours-G}& \text{Ours-L}
			& \text{Schaefer~\cite{schaefer2019ecmr}} & \text{Ours-G}& \text{Ours-L}
			& \text{Schaefer~\cite{schaefer2019ecmr}} & \text{Ours-G}& \text{Ours-L}
			& \text{Schaefer~\cite{schaefer2019ecmr}} & \text{Ours-G} & \text{Ours-L} \\
		\midrule
		2012-01-08 & $\mathrm{0.130}$ &$\mathbf{0.115}$& $\mathrm{0.117}$ & $\mathrm{0.178}$ &$\mathrm{0.146}$&$\mathbf{0.141}$  & $\mathrm{0.663}$ & $\mathbf{0.626}$& $\mathrm{0.633}$  & $\mathrm{0.857}$ & $\mathrm{0.810}$ & $\mathbf{0.808}$ \\
		2012-01-15 & $\mathrm{0.156}$ & $\mathbf{0.146}$&$\mathrm{0.147}$ & $\mathrm{0.225}$ &  $\mathbf{0.204}$& $\mathrm{0.205}$  & $\mathrm{0.760}$ & $\mathbf{0.750}$& $\mathrm{0.753}$  & $\mathrm{0.999}$ & $\mathbf{0.982}$&  $\mathrm{0.989}$\\
		2012-01-22 & $\mathrm{0.172}$ &   $\mathbf{0.149}$& $\mathbf{0.149}$& $\mathrm{0.222}$ &  $\mathrm{0.190}$& $\mathbf{0.185}$  & $\mathrm{0.939}$ &  $\mathbf{0.911}$& $\mathrm{0.912}$  & $\mathrm{1.291}$ & $\mathrm{1.238}$& $\mathbf{1.237}$ \\
		2012-02-02 & $\mathrm{0.155}$ &  $\mathrm{0.136}$& $\mathbf{0.130}$& $\mathrm{0.205}$ &$\mathrm{0.172}$& $\mathbf{0.157}$  & $\mathrm{0.720}$ & $\mathbf{0.699}$& $\mathrm{0.700}$ & $\mathrm{0.975}$ & $\mathrm{0.922}$ & $\mathbf{0.917}$ \\
		2012-02-04 & $\mathrm{0.144}$ & $\mathrm{0.134}$&$\mathbf{0.130}$  & $\mathrm{0.195}$ &$\mathrm{0.167}$& $\mathbf{0.158}$  & $\mathrm{0.684}$ & $\mathrm{0.671}$& $\mathbf{0.670}$ & $\mathrm{0.903}$ & $\mathrm{0.876}$& $\mathbf{0.866}$ \\
		2012-02-05 & $\mathrm{0.148}$ & $\mathbf{0.138}$& $\mathrm{0.145}$ & $\mathrm{0.210}$ & $\mathbf{0.206}$& $\mathrm{0.250}$  & $\mathbf{0.690}$ & $\mathrm{0.694}$& $\mathrm{0.700}$  & $\mathrm{0.947}$ & $\mathbf{0.937}$&  $\mathrm{0.949}$ \\
		2012-02-12 & $\mathrm{0.269}$ & $\mathrm{0.253}$&$\mathbf{0.247}$ & $\mathrm{1.005}$ &  $\mathbf{1.002}$&  $\mathrm{1.003}$ & $\mathrm{0.802}$ &  $\mathrm{0.786}$& $\mathbf{0.778}$  & $\mathrm{1.040}$ &  $\mathrm{1.019}$& $\mathbf{0.997}$ \\
		2012-02-18 & $\mathrm{0.149}$ & $\mathrm{0.131}$& $\mathbf{0.129}$ & $\mathrm{0.221}$ &$\mathrm{0.175}$& $\mathbf{0.161}$  & $\mathrm{0.699}$ & $\mathbf{0.676}$&$\mathrm{0.681}$  & $\mathrm{0.938}$ & $\mathbf{0.905}$& $\mathrm{0.909}$  \\
		2012-02-19 & $\mathrm{0.148}$ &$\mathrm{0.137}$& $\mathbf{0.135}$ & $\mathrm{0.194}$ & $\mathrm{0.183}$&$\mathbf{0.168}$  & $\mathrm{0.704}$ & $\mathbf{0.692}$&  $\mathrm{0.710}$  & $\mathrm{0.944}$ & $\mathbf{0.923}$&  $\mathrm{0.935}$ \\
		2012-03-17 & $\mathrm{0.149}$ &$\mathrm{0.137}$&$\mathbf{0.133}$  & $\mathrm{0.191}$ &$\mathrm{0.174}$& $\mathbf{0.159}$  & $\mathrm{0.830}$ &  $\mathrm{0.798}$& $\mathbf{0.797}$ & $\mathrm{1.062}$ &$\mathrm{1.031}$& $\mathbf{1.022}$  \\
		2012-03-25 & $\mathrm{0.200}$ &  $\mathrm{0.178}$&$\mathbf{0.170}$ & $\mathrm{0.262}$ & $\mathrm{0.235}$&$\mathbf{0.221}$  & $\mathrm{1.418}$ &  $\mathrm{1.379}$& $\mathbf{1.365}$ & $\mathrm{1.836}$ &  $\mathrm{1.789}$& $\mathbf{1.767}$  \\
		2012-03-31 & $\mathrm{0.143}$ &$\mathrm{0.135}$& $\mathbf{0.129}$ & $\mathrm{0.184}$ &$\mathrm{0.176}$ & $\mathbf{0.155}$ & $\mathrm{0.746}$ & $\mathrm{0.729}$& $\mathbf{0.728}$  & $\mathrm{0.973}$ &$\mathrm{0.936}$& $\mathbf{0.932}$  \\
		2012-04-29 & $\mathrm{0.170}$ &$\mathbf{0.154}$& $\mathrm{0.156}$ & $\mathrm{0.251}$ & $\mathbf{0.222}$& $\mathrm{0.227}$    & $\mathrm{0.829}$ & $\mathbf{0.820}$& $\mathrm{0.825}$  & $\mathrm{1.079}$ &$\mathbf{1.069}$& $\mathrm{1.070}$  \\
		2012-05-11 & $\mathrm{0.161}$ & $\mathrm{0.132}$&$\mathbf{0.131}$ & $\mathrm{0.225}$ &$\mathrm{0.163}$ & $\mathbf{0.155}$ & $\mathrm{0.773}$ &  $\mathbf{0.747}$& $\mathrm{0.766}$  & $\mathrm{0.998}$ & $\mathbf{0.965}$& $\mathrm{0.987}$\\
		2012-05-26 & $\mathrm{0.158}$ & $\mathrm{0.142}$& $\mathbf{0.140}$& $\mathrm{0.217}$ &   $\mathrm{0.183}$&$\mathbf{0.165}$  & $\mathrm{0.690}$ &  $\mathbf{0.672}$& $\mathrm{0.681}$  & $\mathrm{0.889}$ & $\mathbf{0.871}$& $\mathrm{0.875}$ \\
		2012-06-15 & $\mathrm{0.180}$ & $\mathbf{0.145}$&$\mathrm{0.147}$ & $\mathrm{0.238}$ & $\mathrm{0.186}$& $\mathbf{0.180}$ & $\mathrm{0.659}$ & $\mathrm{0.646}$& $\mathbf{0.630}$  & $\mathrm{0.879}$ &  $\mathrm{0.874}$& $\mathbf{0.842}$ \\
		2012-08-04 & $\mathrm{0.210}$ & $\mathrm{0.169}$&$\mathbf{0.159}$ & $\mathrm{0.340}$ &$\mathrm{0.230}$& $\mathbf{0.192}$  & $\mathrm{0.884}$ &  $\mathbf{0.843}$&  $\mathrm{0.847}$  & $\mathrm{1.143}$ & $\mathrm{1.093}$& $\mathbf{1.091}$ \\
		2012-08-20 & $\mathrm{0.189}$ & $\mathrm{0.156}$&$\mathbf{0.152}$ & $\mathrm{0.264}$ & $\mathrm{0.207}$ & $\mathbf{0.183}$  & $\mathrm{0.711}$ &  $\mathbf{0.688}$& $\mathrm{0.696}$   & $\mathrm{0.941}$ & $\mathrm{0.906}$& $\mathbf{0.905}$ \\
		2012-09-28 & $\mathrm{0.206}$ &  $\mathrm{0.171}$&$\mathbf{0.155}$& $\mathrm{0.311}$ &  $\mathrm{0.241}$& $\mathbf{0.190}$ & $\mathrm{0.731}$ &  $\mathrm{0.726}$& $\mathbf{0.714}$  & $\mathrm{0.952}$ & $\mathrm{0.949}$&  $\mathbf{0.926}$\\
		2012-10-28 & $\mathrm{0.217}$ &$\mathrm{0.185}$&$\mathbf{0.168}$  & $\mathrm{0.338}$ &$\mathrm{0.281}$& $\mathbf{0.230}$  & $\mathrm{0.693}$ &$\mathrm{0.680}$ & $\mathbf{0.678}$   & $\mathrm{0.919}$ & $\mathrm{0.909}$&  $\mathbf{0.900}$\\
		2012-11-04 & $\mathrm{0.257}$ & $\mathrm{0.208}$&$\mathbf{0.181}$ & $\mathrm{0.456}$ &$\mathrm{0.317}$& $\mathbf{0.227}$  & $\mathrm{0.746}$ & $\mathrm{0.718}$& $\mathbf{0.701}$   & $\mathrm{0.996}$ & $\mathrm{0.973}$& $\mathbf{0.928}$ \\
		2012-11-16 & $\mathrm{0.403}$ &  $\mathrm{0.296}$& $\mathbf{0.251}$& $\mathrm{0.722}$ & $\mathrm{0.435}$&$\mathbf{0.370}$  & $\mathrm{1.467}$ & $\mathrm{1.403}$& $\mathbf{1.378}$ & $\mathrm{2.031}$ &  $\mathrm{1.919}$& $\mathbf{1.895}$  \\
		2012-11-17 & $\mathrm{0.243}$ &$\mathrm{0.201}$& $\mathbf{0.172}$ & $\mathrm{0.377}$ &  $\mathrm{0.323}$& $\mathbf{0.219}$ & $\mathrm{0.686}$ & $\mathrm{0.685}$& $\mathbf{0.677}$  & $\mathrm{0.959}$ & $\mathrm{0.948}$& $\mathbf{0.914}$ \\
		2012-12-01 & $\mathrm{0.266}$ &  $\mathrm{0.226}$&$\mathbf{0.212}$  & $\mathrm{0.492}$ &$\mathbf{0.429}$ & $\mathrm{0.445}$ & $\mathrm{0.674}$ &$\mathrm{0.665}$& $\mathbf{0.647}$   & $\mathrm{0.930}$ & $\mathrm{0.887}$&  $\mathbf{0.854}$ \\
		2013-01-10 & $\mathrm{0.217}$ & $\mathrm{0.187}$&$\mathbf{0.164}$ & $\mathrm{0.278}$ &$\mathrm{0.226}$&$\mathbf{0.190}$   & $\mathrm{0.689}$ & $\mathbf{0.627}$&$\mathrm{0.642}$  & $\mathrm{0.911}$ &$\mathbf{0.806}$&  $\mathrm{0.817}$ \\
		2013-02-23 & $\mathrm{2.470}$  & $\mathrm{0.236}$&$\mathbf{0.207}$ & $\mathrm{5.480}$ & $\mathrm{0.567}$& $\mathbf{0.492}$  & $\mathrm{1.083}$ & $\mathbf{0.592}$& $\mathrm{0.593}$  & $\mathrm{1.769}$ & $\mathrm{0.846}$& $\mathbf{0.845}$ \\
		2013-04-05 & $\mathrm{0.365}$ &$\mathrm{0.295}$& $\mathbf{0.265}$ & $\mathrm{0.920}$ &$\mathrm{0.869}$ & $\mathbf{0.820}$ & $\mathrm{0.654}$ & $\mathbf{0.642}$& $\mathbf{0.642}$ & $\mathrm{1.028}$ &$\mathrm{1.036}$&  $\mathbf{1.016}$ \\
\toprule
		Average	& $\mathrm{0.284}$& $\mathrm{0.174}$& $\mathbf{0.164}$ & $\mathrm{0.526}$&$\mathrm{0.293}$&  $\mathbf{0.268}$ &$\mathrm{0.801}$ & $\mathbf{0.761}$& $\mathbf{0.761}$ & $\mathrm{1.081}$&$\mathrm{1.016}$& $\mathbf{1.007}$ \\
		\bottomrule
	\end{tabular}
}
	\caption{
		Results of our experiments with the NCLT dataset compared to Schaefer~\cite{schaefer2019ecmr}, averaged over ten localization runs per session.
		The variables $\Delta_{\textrm{pos}}$ and $\Delta_{\textrm{ang}}$ denote the mean absolute errors in position and heading, respectively, $\RMSE_{\textrm{pos}}$ and $\RMSE_{\textrm{ang}}$ represent the corresponding root mean squared errors.}
	\label{tab:nclt_results}
\end{table*}

The NCLT dataset contains $27$ sessions with an average length of $5.5\,\mathrm{km}$ and an average duration of $1.3\,\mathrm{h}$ over the course of $15$ months.
The data is recorded at different times over a year, different weather and seasons, including both indoor and outdoor environments, and also lots of dynamic objects. 
The trajectories of different sessions have a large overlap. 
Therefore, it is an ideal dataset for testing long-term localization in urban environments. 

We first build the map following the setup introduced by Schaefer \etal~\cite{schaefer2019ecmr}, which uses the laser scans and the ground-truth poses of the first session. Since in later sessions the robot sometimes moves into unseen places for the first session, we therefore also use those scans whose position is $10\,\mathrm{m}$ away from all previously visited poses to build the map. 
During localization, we use $1000$ particles and use the same initialization as Schaefer \etal~\cite{schaefer2019ecmr} by uniformly sampling positions around the first ground-truth pose within a $2.5\,\mathrm{m}$ circle. The orientations are uniformly sampled from $-5$ to $5$ degrees. 
We resample particles if the number of effective particles is less than $50\,\%$. To get the pose estimation, we use the average poses of the best $10\,\%$ of the particles.

\begin{figure}
	\includegraphics[width=0.95\linewidth]{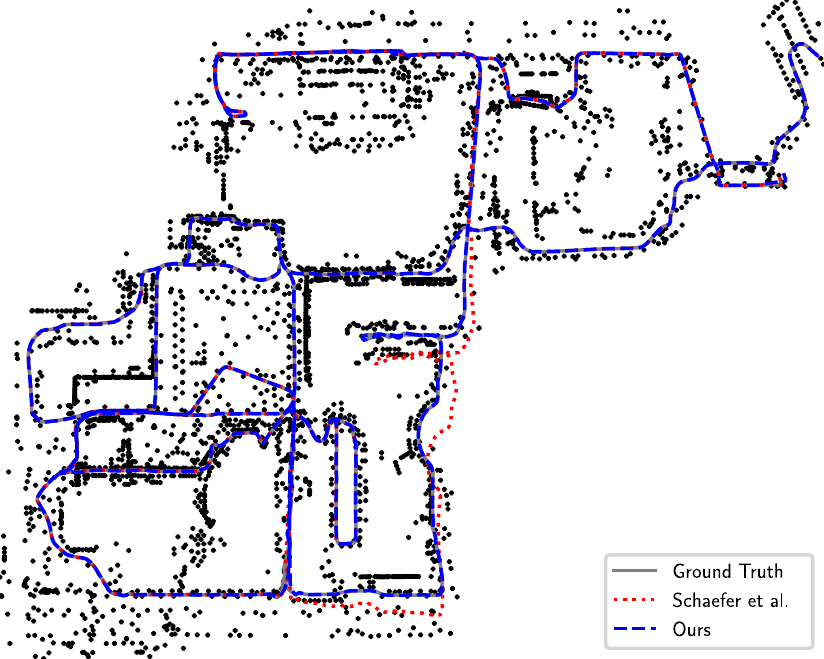}
	\caption{Comparison of localization results of Schaefer \etal~\cite{schaefer2019ecmr} and our method in session 2013-02-23 on NCLT dataset.
	The black dots are the poles on the map. The grey line is the ground-truth trajectory. The blue line is our result and the red one is of the baseline method.
	As can be seen, Schaefer's method loses track of the robot in some places, while our method always tracks the correct robot poses with respect to the ground truth.}
	\label{fig:loc_compare}
  \end{figure}

\tabref{tab:nclt_results} shows the position and orientation
errors for every session. We run the localization $10$ times and compute the average means and RMSEs to the ground-truth trajectory. The results show that both our geometric and learning-based methods surpass Schaefer \etal~\cite{schaefer2019ecmr} in almost all sessions with an average error of $0.174\,\mathrm{m}$ and $0.164\,\mathrm{m}$ respectively.
Besides, in session 2013-02-23, the baseline method fails to localize resulting in an error of $2.470\,\mathrm{m}$, while our method never loses track of the robot position (\figref{fig:loc_compare}).
This is because our pole extractor can robustly extract poles even in an environment where there are fewer poles. Schaefer \etal~\cite{SCHAEFER2021103709} analyze their localization failure in session 2013-02-23 for the reason that the barrels in a construction area are moved a few meters to the right in the later session. As these barrels are detected as poles by their approach, they are built in the map and cause the wrong pole matching to the map in this area. In our pole extraction algorithms, we discard those poles with too large radiuses. Thus, the barrels are not a part of our map and our localization will not be influenced by the movement of these barrels.
Interestingly, our learning-based method improves localization results more than our geometry-based method in most sessions. It may be caused by a more general pole segmentation model trained with pseudo labels generated from different environments.

\begin{table}[t]
  \caption{Localization Results on MulRan Dataset.}
  \centering
  \scalebox{0.85}{
  \begin{tabular}{C{2cm}cccc}
    \toprule
     & Schaefer~\cite{schaefer2019ecmr} & Chen~\cite{chen2021icra}  & Ours-G  & Ours-L\\
	 \midrule
	 $\RMSE_{\mathrm{pos}}$~[\si{m}]  & 1.82 &0.83 & $\mathbf{0.48}$  & 0.49 \\
	 $\RMSE_{\mathrm{ang}}$~[\si{\degree}]  & 0.56 &3.14 & $\mathbf{0.27}$ & 0.28\\
	\bottomrule
  \end{tabular}
}
  \label{tab:mulran_results}
\end{table}

\subsubsection{Localization on the MulRan Dataset}
To further show the generalization ability of our method, we test both our geometric and learning-based methods on the MulRan dataset, which was collected from a different type of LiDAR sensor in a different environment. We use the MulRan dataset KAIST 02 sequence (collected on 2019-08-23) to build the global map and use KAIST 01 sequence (collected on 2019-06-20) for localization.
\tabref{tab:mulran_results} shows the location and yaw angle RMSE
errors on MulRan Dataset. 
As can be seen, our geometric and learning-based methods consistently achieve a better performance than both baseline methods~\cite{schaefer2019ecmr,chen2021icra}.
Note that, we train our pole segmentation only once, and there is no fine-tuning when applying to a new environment.

\subsection{Runtime}

\begin{table}[t]
  \caption{Pole Extraction Runtime Results.}
  \centering
  \small
  \scalebox{0.85}{
  \begin{tabular}{C{2cm}ccc}
    \toprule
     & NCLT & KITTI  & MulRan\\
	 \midrule
	 Ours-G  & 12Hz & 2Hz & 4Hz \\
	 Ours-L  & 17Hz & 16Hz & 16Hz\\
	\bottomrule
  \end{tabular}
}
  \label{tab:speed_results}
\end{table}

This experiment has been conducted to support the claim that our
approach runs online at the sensor frame rate. 
As shown in~\tabref{tab:speed_results}, we compare our method to the baseline method proposed by Schaefer \etal~\cite{schaefer2019ecmr} on three different datasets, including NCLT (session 2012-01-08), KITTI (sequence 09) and MulRan (KAIST 02) datasets. 
As reported in their paper, on NCLT dataset the baseline method takes an average of $1.33\,\mathrm{s}$ for pole extraction on a PC using a GPU.
We tested our geometric method without using a GPU and our method only needs $0.09\,\mathrm{s}$ for pole extraction and all MCL steps take less than $0.1\,\mathrm{s}$ yielding a run time faster than the commonly used LiDAR frame rate of $10\,\mathrm{Hz}$.

The performance of geometry-based pole extractors, both Schaefer's and ours, is influenced by the size of the input data, and it is a trade-off between localization accuracy and speed.
To achieve good localization results for the geometric method, we use the range image size of $32 \times 256$ for NCLT and $64 \times 500$ for KITTI and MulRan, which leads to a decrease in the runtime performance.
However, our learning-based method is not influenced by the size of input data. In our case, we fix the size of network input as $32 \times 256$, and our network always works online with good localization performance with a single GPU, which shows a significant advantage of our learning-based method.
  
\section{Conclusion}
\label{sec:conclusion}

In this paper, we presented a novel range image-based pole extraction approach based on geometric features for online long-term LiDAR localization. Our method exploits range images generated from LiDAR scans. This allows our method to process point cloud data rapidly and run online. We further use the detected poles by our geometric pole extractor as pseudo labels to train a deep neural network for online pole segmentation. Our learning-based pole extractor can generalize to different types of datasets without fine-tuning, despite the environments of different datasets varying a lot. We implemented and evaluated our approach on multiple different datasets and provided comparisons to other existing techniques and supported all claims made in this paper. The experiments suggest that both our geometric and learning-based methods can accurately extract more poles in the environments and achieve better performance in long-term localization tasks than the baseline methods. Moreover, we release our implementation and pole dataset for other researchers to evaluate their algorithms. In the future, we plan to explore the usage of other features such as road markings, curb, and intersection features, to improve the robustness of our method.

\bibliographystyle{cas-model2-names}
\bibliography{glorified,new}

\end{document}

%% file: stachnisslab-latex.tex
\usepackage{graphics}           
\usepackage{times}              
\usepackage{amsmath}            
\usepackage{amssymb}            
\usepackage{graphicx}
\usepackage{algorithm}
\usepackage[noend]{algpseudocode}
\usepackage{booktabs}
\usepackage{color}
\definecolor{instructioncolor}{rgb}{.5,.5,.5}

\usepackage[font=small]{caption}

\def\secref#1{Sec.~\ref{#1}}
\def\figref#1{Fig.~\ref{#1}}
\def\tabref#1{Tab.~\ref{#1}}
\def\eqref#1{Eq.~(\ref{#1})}
\def\algref#1{Alg.~\ref{#1}}

\makeatletter
\usepackage{xspace}
\DeclareRobustCommand\onedot{\futurelet\@let@token\@onedot}
\def\@onedot{\ifx\@let@token.\else.\null\fi\xspace}

\def\etal{{et al}\onedot}
\makeatother

\usepackage{array}
\newcolumntype{L}[1]{>{\raggedright\let\newline\\\arraybackslash\hspace{0pt}}m{#1}}
\newcolumntype{C}[1]{>{\centering\let\newline\\\arraybackslash\hspace{0pt}}m{#1}}
\newcolumntype{R}[1]{>{\raggedleft\let\newline\\\arraybackslash\hspace{0pt}}m{#1}}

%% file: stachnisslab-math.tex
\newcommand{\RR}{\mathbb{R}}

\newcommand{\abs}[1]{|#1|}

\renewcommand{\v}[1]{{\b #1}} 

\newcommand{\q}[1]{{{\bf #1}}}

\newcommand{\cs}[2]{\; ^{#1}\!\,{\mbox{#2}}}










%% file: dong2022ras.bbl
\begin{thebibliography}{47}
\expandafter\ifx\csname natexlab\endcsname\relax\def\natexlab#1{#1}\fi
\providecommand{\href}[2]{#2}
\providecommand{\path}[1]{#1}
\providecommand{\DOIprefix}{doi:}
\providecommand{\ArXivprefix}{arXiv:}
\providecommand{\URLprefix}{}
\providecommand{\Pubmedprefix}{pmid:}
\providecommand{\doi}[1]{\href{http://dx.doi.org/#1}{\path{#1}}}
\providecommand{\Pubmed}[1]{\href{pmid:#1}{\path{#1}}}
\providecommand{\bibinfo}[2]{#2}
\ifx\xfnm\relax \def\xfnm[#1]{\unskip,\space#1}\fi
\bibitem[{Barfoot(2017)}]{10.5555/3165227}
\bibinfo{author}{Barfoot, T.D.}, \bibinfo{year}{2017}.
\newblock \bibinfo{title}{State Estimation for Robotics}.
\newblock \bibinfo{edition}{1st} ed., \bibinfo{publisher}{Cambridge University
  Press}, \bibinfo{address}{USA}.
\bibitem[{Behley et~al.(2019)Behley, Garbade, Milioto, Quenzel, Behnke,
  Stachniss and Gall}]{behley2019iccv}
\bibinfo{author}{Behley, J.}, \bibinfo{author}{Garbade, M.},
  \bibinfo{author}{Milioto, A.}, \bibinfo{author}{Quenzel, J.},
  \bibinfo{author}{Behnke, S.}, \bibinfo{author}{Stachniss, C.},
  \bibinfo{author}{Gall, J.}, \bibinfo{year}{2019}.
\newblock \bibinfo{title}{{SemanticKITTI: A Dataset for Semantic Scene
  Understanding of LiDAR Sequences}}, in: \bibinfo{booktitle}{Proc.~of the
  IEEE/CVF Intl.~Conf.~on Computer Vision (ICCV)}.
\bibitem[{Behley and Stachniss(2018)}]{behley2018rss}
\bibinfo{author}{Behley, J.}, \bibinfo{author}{Stachniss, C.},
  \bibinfo{year}{2018}.
\newblock \bibinfo{title}{{Efficient Surfel-Based SLAM using 3D Laser Range
  Data in Urban Environments}}, in: \bibinfo{booktitle}{Proc.~of Robotics:
  Science and Systems (RSS)}.
\bibitem[{Bennewitz et~al.(2006)Bennewitz, Stachniss, Burgard and
  Behnke}]{bennewitz2006euros}
\bibinfo{author}{Bennewitz, M.}, \bibinfo{author}{Stachniss, C.},
  \bibinfo{author}{Burgard, W.}, \bibinfo{author}{Behnke, S.},
  \bibinfo{year}{2006}.
\newblock \bibinfo{title}{{Metric Localization with Scale-Invariant Visual
  Features using a Single Perspective Camera}}, in:
  \bibinfo{editor}{Christiensen, H.} (Ed.), \bibinfo{booktitle}{European
  Robotics Symposium 2006}, \bibinfo{publisher}{Springer Verlag}. pp.
  \bibinfo{pages}{143--157}.
\bibitem[{Carlevaris-Bianco et~al.(2016)Carlevaris-Bianco, Ushani and
  Eustice}]{carlevaris-bianco2016ijrr}
\bibinfo{author}{Carlevaris-Bianco, N.}, \bibinfo{author}{Ushani, A.},
  \bibinfo{author}{Eustice, R.}, \bibinfo{year}{2016}.
\newblock \bibinfo{title}{{University of Michigan North Campus long-term vision
  and lidar dataset}}.
\newblock \bibinfo{journal}{Intl.~Journal~of Robotics Research (IJRR)}
  \bibinfo{volume}{35}, \bibinfo{pages}{1023--1035}.
\newblock \DOIprefix\doi{10.1177/0278364915614638}.
\bibitem[{Chen et~al.(2021a)Chen, Lu, Li, Liu, Dong, Zhao, Yu and
  Knoll}]{9548787}
\bibinfo{author}{Chen, G.}, \bibinfo{author}{Lu, F.}, \bibinfo{author}{Li, Z.},
  \bibinfo{author}{Liu, Y.}, \bibinfo{author}{Dong, J.}, \bibinfo{author}{Zhao,
  J.}, \bibinfo{author}{Yu, J.}, \bibinfo{author}{Knoll, A.},
  \bibinfo{year}{2021}a.
\newblock \bibinfo{title}{Pole-curb fusion based robust and efficient
  autonomous vehicle localization system with branch-and-bound global
  optimization and local grid map method}.
\newblock \bibinfo{journal}{IEEE Transactions on Vehicular Technology}
  \bibinfo{volume}{70}, \bibinfo{pages}{11283--11294}.
\newblock \DOIprefix\doi{10.1109/TVT.2021.3114825}.
\bibitem[{Chen et~al.(2021b)Chen, L\"abe, Milioto, R\"ohling, Behley and
  Stachniss}]{chen2021auro}
\bibinfo{author}{Chen, X.}, \bibinfo{author}{L\"abe, T.},
  \bibinfo{author}{Milioto, A.}, \bibinfo{author}{R\"ohling, T.},
  \bibinfo{author}{Behley, J.}, \bibinfo{author}{Stachniss, C.},
  \bibinfo{year}{2021}b.
\newblock \bibinfo{title}{{OverlapNet: A Siamese Network for Computing LiDAR
  Scan Similarity with Applications to Loop Closing and Localization}}.
\newblock \bibinfo{journal}{Autonomous Robots}
  \DOIprefix\doi{10.1007/s10514-021-09999-0}.
\bibitem[{Chen et~al.(2020a)Chen, L\"abe, Milioto, R\"ohling, Vysotska, Haag,
  Behley and Stachniss}]{chen2020rss}
\bibinfo{author}{Chen, X.}, \bibinfo{author}{L\"abe, T.},
  \bibinfo{author}{Milioto, A.}, \bibinfo{author}{R\"ohling, T.},
  \bibinfo{author}{Vysotska, O.}, \bibinfo{author}{Haag, A.},
  \bibinfo{author}{Behley, J.}, \bibinfo{author}{Stachniss, C.},
  \bibinfo{year}{2020}a.
\newblock \bibinfo{title}{{OverlapNet: Loop Closing for LiDAR-based SLAM}}, in:
  \bibinfo{booktitle}{Proc.~of Robotics: Science and Systems (RSS)}.
\bibitem[{Chen et~al.(2020b)Chen, L\"abe, Nardi, Behley and
  Stachniss}]{chen2020iros}
\bibinfo{author}{Chen, X.}, \bibinfo{author}{L\"abe, T.},
  \bibinfo{author}{Nardi, L.}, \bibinfo{author}{Behley, J.},
  \bibinfo{author}{Stachniss, C.}, \bibinfo{year}{2020}b.
\newblock \bibinfo{title}{{Learning an Overlap-based Observation Model for 3D
  LiDAR Localization}}, in: \bibinfo{booktitle}{Proc.~of the IEEE/RSJ
  Intl.~Conf.~on Intelligent Robots and Systems (IROS)}.
\bibitem[{Chen et~al.(2021c)Chen, Li, Mersch, Wiesmann, Gall, Behley and
  Stachniss}]{chen2021ral}
\bibinfo{author}{Chen, X.}, \bibinfo{author}{Li, S.}, \bibinfo{author}{Mersch,
  B.}, \bibinfo{author}{Wiesmann, L.}, \bibinfo{author}{Gall, J.},
  \bibinfo{author}{Behley, J.}, \bibinfo{author}{Stachniss, C.},
  \bibinfo{year}{2021}c.
\newblock \bibinfo{title}{{Moving Object Segmentation in 3D LiDAR Data: A
  Learning-based Approach Exploiting Sequential Data}}.
\newblock \bibinfo{journal}{IEEE Robotics and Automation Letters (RA-L)}
  \bibinfo{volume}{6}, \bibinfo{pages}{6529--6536}.
\newblock \DOIprefix\doi{10.1109/LRA.2021.3093567}.
\bibitem[{Chen et~al.(2022)Chen, Mersch, Nunes, Marcuzzi, Vizzo, Behley and
  Stachniss}]{chen2022ral}
\bibinfo{author}{Chen, X.}, \bibinfo{author}{Mersch, B.},
  \bibinfo{author}{Nunes, L.}, \bibinfo{author}{Marcuzzi, R.},
  \bibinfo{author}{Vizzo, I.}, \bibinfo{author}{Behley, J.},
  \bibinfo{author}{Stachniss, C.}, \bibinfo{year}{2022}.
\newblock \bibinfo{title}{{Automatic Labeling to Generate Training Data for
  Online LiDAR-Based Moving Object Segmentation}}.
\newblock \bibinfo{journal}{IEEE Robotics and Automation Letters (RA-L)}
  \bibinfo{volume}{7}, \bibinfo{pages}{6107--6114}.
\newblock \DOIprefix\doi{10.1109/LRA.2022.3166544}.
\bibitem[{Chen et~al.(2019)Chen, Milioto, Palazzolo, Giguère, Behley and
  Stachniss}]{chen2019iros}
\bibinfo{author}{Chen, X.}, \bibinfo{author}{Milioto, A.},
  \bibinfo{author}{Palazzolo, E.}, \bibinfo{author}{Giguère, P.},
  \bibinfo{author}{Behley, J.}, \bibinfo{author}{Stachniss, C.},
  \bibinfo{year}{2019}.
\newblock \bibinfo{title}{{SuMa++: Efficient LiDAR-based Semantic SLAM}}, in:
  \bibinfo{booktitle}{Proc.~of the IEEE/RSJ Intl.~Conf.~on Intelligent Robots
  and Systems (IROS)}.
\bibitem[{Chen et~al.(2021d)Chen, Vizzo, L\"abe, Behley and
  Stachniss}]{chen2021icra}
\bibinfo{author}{Chen, X.}, \bibinfo{author}{Vizzo, I.},
  \bibinfo{author}{L\"abe, T.}, \bibinfo{author}{Behley, J.},
  \bibinfo{author}{Stachniss, C.}, \bibinfo{year}{2021}d.
\newblock \bibinfo{title}{{Range Image-based LiDAR Localization for Autonomous
  Vehicles}}, in: \bibinfo{booktitle}{Proc.~of the IEEE Intl.~Conf.~on Robotics
  \& Automation (ICRA)}.
\bibitem[{Cortinhal et~al.(2020)Cortinhal, Tzelepis and
  Aksoy}]{cortinhal2020iv}
\bibinfo{author}{Cortinhal, T.}, \bibinfo{author}{Tzelepis, G.},
  \bibinfo{author}{Aksoy, E.E.}, \bibinfo{year}{2020}.
\newblock \bibinfo{title}{{SalsaNext: Fast, Uncertainty-Aware Semantic
  Segmentation of LiDAR Point Clouds}}, in: \bibinfo{booktitle}{Proc.~of the
  IEEE Vehicles Symposium (IV)}.
\bibitem[{Cvisic et~al.(2017)Cvisic, Cesic, Markovic and
  Petrovic}]{cvisic2017jfrnewbib}
\bibinfo{author}{Cvisic, I.}, \bibinfo{author}{Cesic, J.},
  \bibinfo{author}{Markovic, I.}, \bibinfo{author}{Petrovic, I.},
  \bibinfo{year}{2017}.
\newblock \bibinfo{title}{{SOFT-SLAM: Computationally Efficient Stereo Visual
  SLAM for Autonomous UAVs}}.
\newblock \bibinfo{journal}{Journal of Field Robotics (JFR)}
  \bibinfo{volume}{35}, \bibinfo{pages}{578--595}.
\bibitem[{Dellaert et~al.(1999)Dellaert, Fox, Burgard and
  Thrun}]{dellaert1999icra}
\bibinfo{author}{Dellaert, F.}, \bibinfo{author}{Fox, D.},
  \bibinfo{author}{Burgard, W.}, \bibinfo{author}{Thrun, S.},
  \bibinfo{year}{1999}.
\newblock \bibinfo{title}{Monte carlo localization for mobile robots}, in:
  \bibinfo{booktitle}{IEEE International Conference on Robotics and Automation
  (ICRA)}.
\bibitem[{Dong et~al.(2021)Dong, Chen and Stachniss}]{dong2021ecmr}
\bibinfo{author}{Dong, H.}, \bibinfo{author}{Chen, X.},
  \bibinfo{author}{Stachniss, C.}, \bibinfo{year}{2021}.
\newblock \bibinfo{title}{{Online Range Image-based Pole Extractor for
  Long-term LiDAR Localization in Urban Environments}}, in:
  \bibinfo{booktitle}{Proc.~of the Europ.~Conf.~on Mobile Robotics (ECMR)}.
\bibitem[{Droeschel and Behnke(2018)}]{droeschel2018icra}
\bibinfo{author}{Droeschel, D.}, \bibinfo{author}{Behnke, S.},
  \bibinfo{year}{2018}.
\newblock \bibinfo{title}{Efficient continuous-time slam for 3d lidar-based
  online mapping}, in: \bibinfo{booktitle}{Proc.~of the IEEE Intl.~Conf.~on
  Robotics \& Automation (ICRA)}.
\bibitem[{Grisetti et~al.(2007)Grisetti, Stachniss and
  Burgard}]{grisetti2007tro}
\bibinfo{author}{Grisetti, G.}, \bibinfo{author}{Stachniss, C.},
  \bibinfo{author}{Burgard, W.}, \bibinfo{year}{2007}.
\newblock \bibinfo{title}{{Improved Techniques for Grid Mapping with
  Rao-Blackwellized Particle Filters}}.
\newblock \bibinfo{journal}{IEEE Trans.~on Robotics (TRO)}
  \bibinfo{volume}{23}, \bibinfo{pages}{34--46}.
\bibitem[{Kim et~al.(2020)Kim, Park, Cho, Jeong and Kim}]{kim2020icra}
\bibinfo{author}{Kim, G.}, \bibinfo{author}{Park, Y.}, \bibinfo{author}{Cho,
  Y.}, \bibinfo{author}{Jeong, J.}, \bibinfo{author}{Kim, A.},
  \bibinfo{year}{2020}.
\newblock \bibinfo{title}{Mulran: Multimodal range dataset for urban place
  recognition}, in: \bibinfo{booktitle}{Proc.~of the IEEE Intl.~Conf.~on
  Robotics \& Automation (ICRA)}.
\bibitem[{K{\"u}mmerle et~al.(2019)K{\"u}mmerle, Sons, Poggenhans, K{\"u}hner,
  Lauer and Stiller}]{kummerle2019icra}
\bibinfo{author}{K{\"u}mmerle, J.}, \bibinfo{author}{Sons, M.},
  \bibinfo{author}{Poggenhans, F.}, \bibinfo{author}{K{\"u}hner, T.},
  \bibinfo{author}{Lauer, M.}, \bibinfo{author}{Stiller, C.},
  \bibinfo{year}{2019}.
\newblock \bibinfo{title}{Accurate and efficient self-localization on roads
  using basic geometric primitives}, in: \bibinfo{booktitle}{Proc.~of the IEEE
  Intl.~Conf.~on Robotics \& Automation (ICRA)}.
\bibitem[{K{\"u}mmerle et~al.(2014)K{\"u}mmerle, Ruhnke, Steder, Stachniss and
  Burgard}]{kummerle14jfr}
\bibinfo{author}{K{\"u}mmerle, R.}, \bibinfo{author}{Ruhnke, M.},
  \bibinfo{author}{Steder, B.}, \bibinfo{author}{Stachniss, C.},
  \bibinfo{author}{Burgard, W.}, \bibinfo{year}{2014}.
\newblock \bibinfo{title}{Autonomous robot navigation in highly populated
  pedestrian zones}.
\newblock \bibinfo{journal}{Journal of Field Robotics (JFR)}
  \DOIprefix\doi{10.1002/rob.21534}.
\bibitem[{Li et~al.(2022)Li, Chen, Liu, Dai, Stachniss and Gall}]{li2022ral}
\bibinfo{author}{Li, S.}, \bibinfo{author}{Chen, X.}, \bibinfo{author}{Liu,
  Y.}, \bibinfo{author}{Dai, D.}, \bibinfo{author}{Stachniss, C.},
  \bibinfo{author}{Gall, J.}, \bibinfo{year}{2022}.
\newblock \bibinfo{title}{{Multi-scale Interaction for Real-time LiDAR Data
  Segmentation on an Embedded Platform}}.
\newblock \bibinfo{journal}{IEEE Robotics and Automation Letters (RA-L)}
  \bibinfo{volume}{7}, \bibinfo{pages}{738--745}.
\newblock \DOIprefix\doi{10.1109/LRA.2021.3132059}.
\bibitem[{Ma et~al.(2019)Ma, Tartavull, B{\^a}rsan, Wang, Bai, Mattyus,
  Homayounfar, Lakshmikanth, Pokrovsky and Urtasun}]{ma2019iros}
\bibinfo{author}{Ma, W.}, \bibinfo{author}{Tartavull, I.},
  \bibinfo{author}{B{\^a}rsan, I.A.}, \bibinfo{author}{Wang, S.},
  \bibinfo{author}{Bai, M.}, \bibinfo{author}{Mattyus, G.},
  \bibinfo{author}{Homayounfar, N.}, \bibinfo{author}{Lakshmikanth, S.K.},
  \bibinfo{author}{Pokrovsky, A.}, \bibinfo{author}{Urtasun, R.},
  \bibinfo{year}{2019}.
\newblock \bibinfo{title}{{Exploiting Sparse Semantic HD Maps for Self-Driving
  Vehicle Localization}}, in: \bibinfo{booktitle}{Proc.~of the IEEE/RSJ
  Intl.~Conf.~on Intelligent Robots and Systems (IROS)}.
\bibitem[{Maddern et~al.(2017)Maddern, Pascoe, Linegar and Newman}]{RobotCar}
\bibinfo{author}{Maddern, W.}, \bibinfo{author}{Pascoe, G.},
  \bibinfo{author}{Linegar, C.}, \bibinfo{author}{Newman, P.},
  \bibinfo{year}{2017}.
\newblock \bibinfo{title}{1 year, 1000 km: The oxford robotcar dataset}.
\newblock \bibinfo{journal}{The International Journal of Robotics Research}
  \bibinfo{volume}{36}, \bibinfo{pages}{3--15}.
\newblock \DOIprefix\doi{10.1177/0278364916679498}.
\bibitem[{Milioto and Stachniss(2019)}]{milioto2019icra}
\bibinfo{author}{Milioto, A.}, \bibinfo{author}{Stachniss, C.},
  \bibinfo{year}{2019}.
\newblock \bibinfo{title}{{Bonnet: An Open-Source Training and Deployment
  Framework for Semantic Segmentation in Robotics using CNNs}}, in:
  \bibinfo{booktitle}{Proc.~of the IEEE Intl.~Conf.~on Robotics \& Automation
  (ICRA)}.
\bibitem[{Milioto et~al.(2019)Milioto, Vizzo, Behley and
  Stachniss}]{milioto2019iros}
\bibinfo{author}{Milioto, A.}, \bibinfo{author}{Vizzo, I.},
  \bibinfo{author}{Behley, J.}, \bibinfo{author}{Stachniss, C.},
  \bibinfo{year}{2019}.
\newblock \bibinfo{title}{{RangeNet++: Fast and Accurate LiDAR Semantic
  Segmentation}}, in: \bibinfo{booktitle}{Proceedings of the IEEE/RSJ Int.
  Conf. on Intelligent Robots and Systems (IROS)}.
\bibitem[{Mur-Artal et~al.(2015)Mur-Artal, Montiel and
  Tardos}]{mur-artal2015tro}
\bibinfo{author}{Mur-Artal, R.}, \bibinfo{author}{Montiel, J.},
  \bibinfo{author}{Tardos, J.D.}, \bibinfo{year}{2015}.
\newblock \bibinfo{title}{{ORB-SLAM: a versatile and accurate monocular SLAM
  system}}.
\newblock \bibinfo{journal}{IEEE Trans.~on Robotics (TRO)}
  \bibinfo{volume}{31}, \bibinfo{pages}{1147--1163}.
\bibitem[{Plachetka et~al.(2021)Plachetka, Fricke, Klingner and
  Fingscheidt}]{9564759}
\bibinfo{author}{Plachetka, C.}, \bibinfo{author}{Fricke, J.},
  \bibinfo{author}{Klingner, M.}, \bibinfo{author}{Fingscheidt, T.},
  \bibinfo{year}{2021}.
\newblock \bibinfo{title}{Dnn-based recognition of pole-like objects in lidar
  point clouds}, in: \bibinfo{booktitle}{2021 IEEE International Intelligent
  Transportation Systems Conference (ITSC)}, pp. \bibinfo{pages}{2889--2896}.
\newblock \DOIprefix\doi{10.1109/ITSC48978.2021.9564759}.
\bibitem[{Schaefer et~al.(2019)Schaefer, B{\"u}scher, Vertens, Luft and
  Burgard}]{schaefer2019ecmr}
\bibinfo{author}{Schaefer, A.}, \bibinfo{author}{B{\"u}scher, D.},
  \bibinfo{author}{Vertens, J.}, \bibinfo{author}{Luft, L.},
  \bibinfo{author}{Burgard, W.}, \bibinfo{year}{2019}.
\newblock \bibinfo{title}{{Long-term urban vehicle localization using pole
  landmarks extracted from 3-D lidar scans}}, in: \bibinfo{booktitle}{Proc.~of
  the Europ.~Conf.~on Mobile Robotics (ECMR)}, pp. \bibinfo{pages}{1--7}.
\bibitem[{Schaefer et~al.(2021)Schaefer, Büscher, Vertens, Luft and
  Burgard}]{SCHAEFER2021103709}
\bibinfo{author}{Schaefer, A.}, \bibinfo{author}{Büscher, D.},
  \bibinfo{author}{Vertens, J.}, \bibinfo{author}{Luft, L.},
  \bibinfo{author}{Burgard, W.}, \bibinfo{year}{2021}.
\newblock \bibinfo{title}{Long-term vehicle localization in urban environments
  based on pole landmarks extracted from 3-d lidar scans}.
\newblock \bibinfo{journal}{Robotics and Autonomous Systems}
  \bibinfo{volume}{136}.
\newblock \DOIprefix\doi{https://doi.org/10.1016/j.robot.2020.103709}.
\bibitem[{Sefati et~al.(2017)Sefati, Daum, Sondermann, Kreisk{\"o}ther and
  Kampker}]{sefati2017iv}
\bibinfo{author}{Sefati, M.}, \bibinfo{author}{Daum, M.},
  \bibinfo{author}{Sondermann, B.}, \bibinfo{author}{Kreisk{\"o}ther, K.D.},
  \bibinfo{author}{Kampker, A.}, \bibinfo{year}{2017}.
\newblock \bibinfo{title}{Improving vehicle localization using semantic and
  pole-like landmarks}, in: \bibinfo{booktitle}{Proc.~of the IEEE Vehicles
  Symposium (IV)}.
\bibitem[{Shi et~al.(2021)Shi, Chen, Huang, Xiao, Lu and
  Stachniss}]{shi2021ral}
\bibinfo{author}{Shi, C.}, \bibinfo{author}{Chen, X.}, \bibinfo{author}{Huang,
  K.}, \bibinfo{author}{Xiao, J.}, \bibinfo{author}{Lu, H.},
  \bibinfo{author}{Stachniss, C.}, \bibinfo{year}{2021}.
\newblock \bibinfo{title}{{Keypoint Matching for Point Cloud Registration using
  Multiplex Dynamic Graph Attention Networks}}.
\newblock \bibinfo{journal}{IEEE Robotics and Automation Letters (RA-L)}
  \bibinfo{volume}{6}, \bibinfo{pages}{8221--8228}.
\newblock \DOIprefix\doi{10.1109/LRA.2021.3097275}.
\bibitem[{Shi et~al.(2018)Shi, Kang, Lin, Liu and Chen}]{rs10121891}
\bibinfo{author}{Shi, Z.}, \bibinfo{author}{Kang, Z.}, \bibinfo{author}{Lin,
  Y.}, \bibinfo{author}{Liu, Y.}, \bibinfo{author}{Chen, W.},
  \bibinfo{year}{2018}.
\newblock \bibinfo{title}{Automatic recognition of pole-like objects from
  mobile laser scanning point clouds}.
\newblock \bibinfo{journal}{Remote Sensing} \bibinfo{volume}{10}.
\newblock \DOIprefix\doi{10.3390/rs10121891}.
\bibitem[{Spangenberg et~al.(2016)Spangenberg, Goehring and
  Rojas}]{spangenberg2016iros}
\bibinfo{author}{Spangenberg, R.}, \bibinfo{author}{Goehring, D.},
  \bibinfo{author}{Rojas, R.}, \bibinfo{year}{2016}.
\newblock \bibinfo{title}{{Pole-Based Localization for Autonomous Vehicles in
  Urban Scenarios}}, in: \bibinfo{booktitle}{Proc.~of the IEEE/RSJ
  Intl.~Conf.~on Intelligent Robots and Systems (IROS)}.
\bibitem[{Sun et~al.(2020)Sun, Adolfsson, Magnusson, Andreasson, Posner and
  Duckett}]{sun2020icra}
\bibinfo{author}{Sun, L.}, \bibinfo{author}{Adolfsson, D.},
  \bibinfo{author}{Magnusson, M.}, \bibinfo{author}{Andreasson, H.},
  \bibinfo{author}{Posner, I.}, \bibinfo{author}{Duckett, T.},
  \bibinfo{year}{2020}.
\newblock \bibinfo{title}{{Localising Faster: Efficient and precise lidar-based
  robot localisation in large-scale environments}}, in:
  \bibinfo{booktitle}{Proc.~of the IEEE Intl.~Conf.~on Robotics \& Automation
  (ICRA)}.
\bibitem[{Thrun et~al.(2005)Thrun, Burgard and Fox}]{thrun2005probrobbook}
\bibinfo{author}{Thrun, S.}, \bibinfo{author}{Burgard, W.},
  \bibinfo{author}{Fox, D.}, \bibinfo{year}{2005}.
\newblock \bibinfo{title}{{Probabilistic Robotics}}.
\newblock \bibinfo{publisher}{MIT Press}.
\bibitem[{Tinchev et~al.(2019)Tinchev, Penate-Sanchez and
  Fallon}]{tinchev2019ral}
\bibinfo{author}{Tinchev, G.}, \bibinfo{author}{Penate-Sanchez, A.},
  \bibinfo{author}{Fallon, M.}, \bibinfo{year}{2019}.
\newblock \bibinfo{title}{{Learning to see the wood for the trees: Deep laser
  localization in urban and natural environments on a CPU}}.
\newblock \bibinfo{journal}{IEEE Robotics and Automation Letters (RA-L)}
  \bibinfo{volume}{4}, \bibinfo{pages}{1327--1334}.
\bibitem[{Toft et~al.(2020)Toft, Maddern, Torii, Hammarstrand, Stenborg,
  Safari, Okutomi, Pollefeys, Sivic, Pajdla, Kahl and Sattler}]{toft2020tpami}
\bibinfo{author}{Toft, C.}, \bibinfo{author}{Maddern, W.},
  \bibinfo{author}{Torii, A.}, \bibinfo{author}{Hammarstrand, L.},
  \bibinfo{author}{Stenborg, E.}, \bibinfo{author}{Safari, D.},
  \bibinfo{author}{Okutomi, M.}, \bibinfo{author}{Pollefeys, M.},
  \bibinfo{author}{Sivic, J.}, \bibinfo{author}{Pajdla, T.},
  \bibinfo{author}{Kahl, F.}, \bibinfo{author}{Sattler, T.},
  \bibinfo{year}{2020}.
\newblock \bibinfo{title}{Long-term visual localization revisited}.
\newblock \bibinfo{journal}{IEEE Trans.~on Pattern Analalysis and Machine
  Intelligence (TPAMI)} \DOIprefix\doi{10.1109/TPAMI.2020.3032010}.
\bibitem[{Trahanias et~al.(2005)Trahanias, Burgard, Argyros, H\"{a}hnel,
  Baltzakis, Pfaff and Stachniss}]{trahanias2005ram}
\bibinfo{author}{Trahanias, P.}, \bibinfo{author}{Burgard, W.},
  \bibinfo{author}{Argyros, A.}, \bibinfo{author}{H\"{a}hnel, D.},
  \bibinfo{author}{Baltzakis, H.}, \bibinfo{author}{Pfaff, P.},
  \bibinfo{author}{Stachniss, C.}, \bibinfo{year}{2005}.
\newblock \bibinfo{title}{{TOURBOT and WebFAIR: Web-Operated Mobile Robots for
  Tele-Presence in Populated Exhibitions}}.
\newblock \bibinfo{journal}{IEEE Robotics and Automation Magazine (RAM)}
  \bibinfo{volume}{12}, \bibinfo{pages}{77--89}.
\bibitem[{Vizzo et~al.(2021)Vizzo, Chen, Chebrolu, Behley and
  Stachniss}]{vizzo2021icra}
\bibinfo{author}{Vizzo, I.}, \bibinfo{author}{Chen, X.},
  \bibinfo{author}{Chebrolu, N.}, \bibinfo{author}{Behley, J.},
  \bibinfo{author}{Stachniss, C.}, \bibinfo{year}{2021}.
\newblock \bibinfo{title}{{Poisson Surface Reconstruction for LiDAR Odometry
  and Mapping}}, in: \bibinfo{booktitle}{Proc.~of the IEEE Intl.~Conf.~on
  Robotics \& Automation (ICRA)}.
\bibitem[{Weng et~al.(2018)Weng, Yang, Guo, Wang and Wang}]{weng2018rcar}
\bibinfo{author}{Weng, L.}, \bibinfo{author}{Yang, M.}, \bibinfo{author}{Guo,
  L.}, \bibinfo{author}{Wang, B.}, \bibinfo{author}{Wang, C.},
  \bibinfo{year}{2018}.
\newblock \bibinfo{title}{Pole-based real-time localization for autonomous
  driving in congested urban scenarios}, in: \bibinfo{booktitle}{Proc.~of the
  Intl.~Conf.~on Real-time Computing and Robotics (RCAR)}.
\bibitem[{Weng et~al.(2016)Weng, Li, Chen and Wang}]{weng2016road}
\bibinfo{author}{Weng, S.}, \bibinfo{author}{Li, J.}, \bibinfo{author}{Chen,
  Y.}, \bibinfo{author}{Wang, C.}, \bibinfo{year}{2016}.
\newblock \bibinfo{title}{{Road traffic sign detection and classification from
  mobile LiDAR point clouds}}, in: \bibinfo{booktitle}{Proc.~of the
  Intl.~Conf.~on Computer Vision in Remote Sensing}.
\bibitem[{Wiesmann et~al.(2021)Wiesmann, Milioto, Chen, Stachniss and
  Behley}]{wiesmann2021ral}
\bibinfo{author}{Wiesmann, L.}, \bibinfo{author}{Milioto, A.},
  \bibinfo{author}{Chen, X.}, \bibinfo{author}{Stachniss, C.},
  \bibinfo{author}{Behley, J.}, \bibinfo{year}{2021}.
\newblock \bibinfo{title}{{Deep Compression for Dense Point Cloud Maps}}.
\newblock \bibinfo{journal}{IEEE Robotics and Automation Letters (RA-L)}
  \bibinfo{volume}{6}, \bibinfo{pages}{2060--2067}.
\newblock \DOIprefix\doi{10.1109/LRA.2021.3059633}.
\bibitem[{Wilbers(2021)}]{wilbers2021phd}
\bibinfo{author}{Wilbers, D.}, \bibinfo{year}{2021}.
\newblock \bibinfo{title}{Graph-based Sliding Window Localization and Map
  Refinement for Automated Vehicles}.
\newblock Ph.D. thesis. Rheinische Friedrich-Wilhelms-Universität Bonn.
\bibitem[{Zhou et~al.(2018)Zhou, Park and Koltun}]{zhou2018arXiv}
\bibinfo{author}{Zhou, Q.}, \bibinfo{author}{Park, J.},
  \bibinfo{author}{Koltun, V.}, \bibinfo{year}{2018}.
\newblock \bibinfo{title}{{Open3D}: {A} modern library for {3D} data
  processing}.
\newblock \bibinfo{journal}{arXiv:1801.09847} .
\bibitem[{Zimmerman et~al.(2022)Zimmerman, Wiesmann, Guadagnino, Läbe, Behley
  and Stachniss}]{zimmerman2022iros}
\bibinfo{author}{Zimmerman, N.}, \bibinfo{author}{Wiesmann, L.},
  \bibinfo{author}{Guadagnino, T.}, \bibinfo{author}{Läbe, T.},
  \bibinfo{author}{Behley, J.}, \bibinfo{author}{Stachniss, C.},
  \bibinfo{year}{2022}.
\newblock \bibinfo{title}{{Robust Onboard Localization in Changing Environments
  Exploiting Text Spotting}}, in: \bibinfo{booktitle}{Proc.~of the IEEE/RSJ
  Intl.~Conf.~on Intelligent Robots and Systems (IROS)}.

\end{thebibliography}
